\pdfoutput=1

\documentclass[11pt]{article}
\usepackage[table, dvipsnames]{xcolor}
\usepackage[preprint]{acl}

\definecolor{groupgray}{gray}{0.93}

\usepackage{times}
\usepackage{latexsym}

\usepackage[T1]{fontenc}

\usepackage[utf8]{inputenc}

\usepackage{microtype}

\usepackage{inconsolata}

\usepackage{hyperref}
\usepackage{url}

\usepackage[most]{tcolorbox}

\colorlet{lightSalmon}{Salmon!80}

\usepackage{graphicx}
\usepackage{subfigure}
\usepackage{wrapfig}

\newcommand{\mybox}[4]{
    \begin{figure}[h]
        \centering
    \begin{tikzpicture}
        \node[anchor=text,text width=\columnwidth-0.45cm, draw, rounded corners, line width=1pt, fill=#3, inner sep=2.5mm, align=justify] (big) {\\#4};
        \node[draw, rounded corners, line width=1pt, fill=#2, anchor=west, xshift=5mm] (small) at (big.north west) {#1};
    \end{tikzpicture}
    \end{figure}
}

\usepackage{amsmath}
\usepackage{amstext}
\usepackage{amsfonts}
\usepackage{bm}

\usepackage{bbm}
\usepackage{tabularx}
\usepackage{multirow}
\usepackage{booktabs}
\usepackage{array}
\usepackage{caption}
\usepackage{multirow}
\usepackage{booktabs}
\usepackage{array}
\usepackage{caption}
\usepackage{color}
\usepackage{colortbl}
\usepackage{tablefootnote}
\usepackage{adjustbox}

\usepackage{caption}
\usepackage{algorithm}
\usepackage{algpseudocode}

\algnewcommand{\LineComment}[1]{\Statex ~~~~~~\textsc{//}~\textit{#1}}

\usepackage{enumitem}
\setenumerate[1]{itemsep=0pt,partopsep=0pt,parsep=\parskip,topsep=5pt}
\setitemize[1]{itemsep=0pt,partopsep=0pt,parsep=\parskip,topsep=5pt}
\setdescription{itemsep=0pt,partopsep=0pt,parsep=\parskip,topsep=5pt}

\usepackage{soul}

\usepackage{tikz}
\usepackage[edges]{forest}
\definecolor{hidden-draw}{RGB}{64,101,149}
\definecolor{hidden-pink}{RGB}{231,239,250}

\usepackage[mathscr]{euscript}

\usepackage{amssymb}  
\usepackage{pifont}

\usepackage{xspace}

\newcommand{\method}{\texttt{Cot Compress}\xspace}

\usepackage{ulem}

\usepackage{listings}
\usepackage{xcolor}

\lstdefinelanguage{json}{
  basicstyle=\ttfamily\small,
  numbers=none,
  breaklines=true,
  showstringspaces=false,
  string=[s]{"}{"},
  morecomment=[l]{//},
  morecomment=[s]{/*}{*/},
  morekeywords={true,false,null},
}

\lstnewenvironment{JSON}
  {\lstset{language=json}}
  {}

\title{When Compression Helps and When It Hurts: \\Condition-Aware Analysis of Chain-of-Thought Distillation
}

\author{
    Siyang Lyu\textsuperscript{\rm 1,2}\thanks{Equal Contributions.},
    Xinghao Chen\textsuperscript{\rm 1,3}\footnotemark[1],
    Zhijing Sun\textsuperscript{\rm 1,2},\\
    \textbf{Tong Liu}\textsuperscript{\rm 4},
    \textbf{Dawei Zhu}\textsuperscript{\rm 5},
    \textbf{Xiaoyu Shen}\textsuperscript{\rm 1}\thanks{Corresponding authors.}
\\
  \textsuperscript{\rm 1}Zhejiang Key Laboratory of Industrial Intelligence and Digital Twin, EIT, Ningbo\\
  \textsuperscript{\rm 2}Viterbi School of Engineering, University of Southern California\\
  \textsuperscript{\rm 3} The Hong Kong Polytechnic University
  \textsuperscript{\rm 4} LMU Munich\\
  \textsuperscript{\rm 5} Saarland University, Saarland Informatics Campus\\
  {\tt siyangly@usc.edu ~ xyshen@eitech.edu.cn}
}

\begin{document}
 \maketitle
\begin{abstract}
Chain-of-Thought (CoT) distillation transfers multi-step reasoning from large reasoning models to smaller students, but verbose teacher traces inflate both training and inference cost. Existing CoT compression methods fall into two families, selective pruning and generative rewriting, yet prior studies have left key factors entangled: granularity is confounded with importance criteria in pruning, restructuring level is rarely isolated in rewriting, and compression budgets are not systematically evaluated across domains or regimes. We recast CoT compression along three dimensions: importance criterion, restructuring level, and compression budget. Sweeping these across two model families, Math and General domains, and Long-/Short-CoT regimes, we find that (i) importance criterion utility is strictly governed by granularity: step-level criteria converge on a shared reasoning backbone, while token-level pruning requires symbol-aware signals to preserve the logical core; (ii) restructuring level inverts across domains: Math degrades monotonically with structural disruption, while aggressive rewriting acts as a denoiser on General tasks; (iii) training-time compression does not necessarily translate to inference-time savings: Long-CoT students retain verbose habits despite concise supervision, making the training ratio an optimistic lower bound on deployment cost. These findings yield condition-aware guidelines for matching compression to deployment context.\footnote{The code will be released at \url{https://github.com/EIT-NLP/CoT-Distillation-Compression}.}
\end{abstract}

\newcommand{\researchquestion}[1]{%
    \par\noindent 
    \fbox{%
        \parbox{\dimexpr\linewidth-2\fboxsep-2\fboxrule\relax}{%
            \centering\itshape #1%
        }%
    }\par\addvspace{\baselineskip}%
}

\begin{figure}[t]
\centering
\includegraphics[width=1\linewidth]{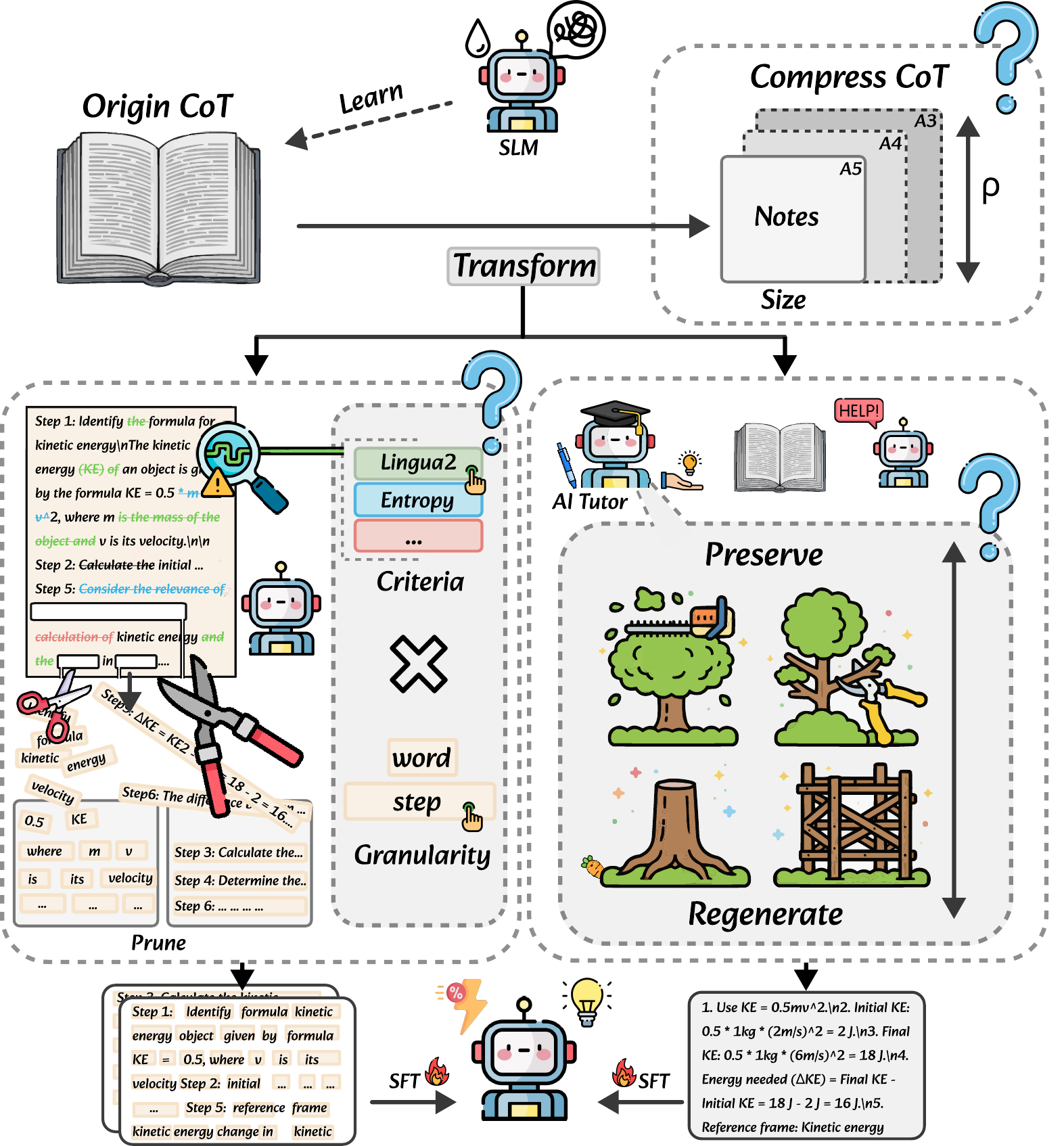}
\caption{The CoT Compression Pipeline. To reduce computational costs, verbose reasoning is compressed by pruning low-value words or regenerating core logic, producing concise training data for efficient SFT.}
\vspace{-12pt}
\label{fig:main}
\end{figure}

\section{Introduction}

Large reasoning models (LRMs,~\citealt{chen-etal-2025-towards-reasoning-era}) have recently achieved strong performance on complex reasoning tasks~\citep{deepseekai2025deepseekr1incentivizingreasoningcapability, openai2024_learning_reason_llms}.
Because directly deploying these models is costly, a common strategy is to distill their reasoning traces into smaller student models~\citep{huang2024o1replicationjourney, ye2025limoreasoning}.
However, these traces are often verbose, containing redundant verification, speculative detours, and repeated reformulations~\citep{chen2025think23overthinkingo1like, wang2025thoughtsplaceunderthinking, zhen2025reconsideringoverthinking}. Such verbosity increases training-time token consumption and may also teach students to generate unnecessarily long reasoning traces at inference time. CoT compression therefore aims to shorten reasoning supervision while preserving the information needed for effective distillation.

Existing CoT compression methods generally fall into two families. \textit{Selective pruning} deletes parts of the original trace using saliency, uncertainty, surprisal, or related importance scores~\citep{pan:2024llmlingua2, xia2025tokenskip, li2026stepentropy, zhao2025pruneonlogic, xu2025astarthought}. \textit{Generative rewriting} produces a new shorter rationale through paraphrasing, step fusion, or structured regeneration~\citep{yu2025lsmixturesft, kang2025c3ot, wu2025dap, jiang2026drp}. While both families successfully reduce cot length, current research still lacks a unified understanding of \textit{what underlying factors actually drive effective compression}. In particular, three critical issues remain unresolved.

First, for \textit{selective pruning} methods, the role of \textit{granularity} is insufficiently understood. Existing approaches typically operate either at the step level~\citep{li2026stepentropy, zhao2025pruneonlogic} or at the token level~\citep{xia2025tokenskip}, but granularity is often entangled with the importance criterion itself. As a result, it remains unclear whether performance differences arise from the scoring signal or from the level at which compression is applied.

Second, for \textit{generative rewriting} methods, the role of \textit{restructuring level} is rarely studied in isolation. Existing rewriting approaches implicitly introduce different degrees of structural transformation~\citep{yu2025lsmixturesft, kang2025c3ot, wu2025dap}, ranging from light paraphrasing to aggressive regeneration of reasoning traces. However, prior work neither isolates restructuring level as an independent variable nor controls it against compression ratio, making it difficult to determine whether performance differences stem from information preservation or structural disruption.

Third, across both pruning and rewriting methods, the role of \textit{compression budget} lacks systematic evaluation. Existing studies typically report token reduction and downstream accuracy under a single task domain or compression setting, without examining how compression behavior changes across domains, reasoning regimes, or compression intensities~\citep{jiang2026drp, wu2025dap, wang2025r1compress}.

Collectively, these gaps reduce CoT compression to trial-and-error tuning rather than principled design.
In this work, we address these limitations by reframing CoT compression as a controlled design problem along three dimensions~(Figure~\ref{fig:main}):
\textit{importance criterion}, which determines what reasoning content is retained;
\textit{restructuring level}, which characterizes how strongly the original reasoning structure is modified; and
\textit{compression budget}, which distinguishes the target compression ratio, the realized training ratio, and the actual inference-time token ratio during deployment.
We systematically study these dimensions across two model families, two task domains, and both Short-CoT and Long-CoT supervision regimes.

Our analysis reveals the following principal findings. \textbf{(i)} Importance criterion utility is strictly governed by granularity: at the step level, distinct criteria converge on a shared reasoning backbone, rendering criterion choice largely irrelevant; at the token level, this robustness collapses, and only symbol-aware signals preserve the logical core under aggressive pruning. \textbf{(ii)} The effect of restructuring level inverts across domains: on Math, accuracy degrades monotonically as structural disruption increases, leaving structure-preserving methods as the only reliable choice; on General tasks, aggressive rewriting acts as a denoiser, matching or exceeding structure-preserving baselines. \textbf{(iii)} Training-time compression does not necessarily translate into inference-time efficiency. Under Short-CoT supervision, students learn genuinely concise reasoning patterns and achieve real deployment savings. Under Long-CoT supervision, however, students retain verbose reasoning behaviors even when trained on compressed traces, causing the realized inference cost to remain substantially higher than the nominal training compression ratio.

These findings lead to a practical guideline for choosing CoT compression strategies.
For symbolic or math-heavy tasks, compression should preserve the original reasoning structure, and token-level pruning should use symbol-aware criteria that protect constraints, operators, and intermediate values.
For general-domain reasoning, more aggressive semantic fusion or framework-level rewriting can be beneficial because it can remove noise without destroying a unique derivation path.
When deployment efficiency is the goal, practitioners should measure the student's actual inference length rather than relying only on compressed training-token counts.
Overall, our study shows that effective CoT compression is not a universal token-reduction recipe, but a set of condition-aware choices that must be matched to the structure of the task and the intended deployment setting.

\section{Related Works}
\label{sec:Related Works}

\paragraph{Overthinking in LRMs} 
Recent LRMs, like OpenAI’s o1 ~\citep{openai2024_learning_reason_llms}
and DeepSeek-R1 ~\citep{deepseekai2025deepseekr1incentivizingreasoningcapability}, have demonstrated strong complex-task performance. Distilling their reasoning traces enables smaller models to acquire reasoning abilities~\citep{huang2024o1replicationjourney}. However, distillation also transfers the teacher’s overthinking behavior~\citep{chen2025think23overthinkingo1like}—redundant verification, speculative detours, and repetitive restates that inflate both training-time token budgets and inference-time latency without significant scaling benefits in student performance~\citep{wang2025harnessingreasoningeconomysurvey, chen-etal-2025-unveiling-key, wang2025thoughtsplaceunderthinking}. 
Motivated by this, recent work on efficient CoT explores ways to retain the utility of explicit reasoning while reducing its cost~\citep{sui2025stopoverthinkingsurveyefficient}.
A complementary line of work tackles overthinking from a model-architecture perspective, examining the effectiveness of pruning strategies on reasoning-augmented LRMs~\citep{ding2026llmslrmsrethinkingpruning} and developing dynamic layer pruning that adaptively allocates computation per token~\citep{zhao2025skipgpt}.
In parallel, a growing body of work studies \textbf{CoT compression} by pruning intermediate steps or compressing teacher traces via summarization and rewriting, thereby obtaining cheaper supervision for student training~\citep{zhu2025conciseadaptivethinkinglarge, sui2025stopoverthinkingsurveyefficient}.


\paragraph{Trace-level CoT Compression: Selective Pruning and Generative Rewriting}
One thread, \textit{selective preservation and rationale pruning}, retains critical trace segments: saliency-based estimators isolate informative spans to remove fluff ~\citep{pan:2024llmlingua2, xia2025tokenskip}; information-theoretic signals, uncertainty or surprisal, identify essential reasoning blocks ~\citep{li2026stepentropy, cui2025stepwiseperplexity, xu2025astarthought}; and dependency-aware frameworks preserve logically indispensable components by modeling inter-step relations through graphs or heuristic rules~\citep{zhao2025pruneonlogic, jia2025epic}. 
A complementary thread, \textit{generative refinement and structural re-organization}, paraphrases and condenses teacher traces across a transformation spectrum—from skeleton-preserving polishing ~\citep{yu2025lsmixturesft}, to step-merging into abstract statements ~\citep{jiang2026drp, kang2025c3ot}, and intent-guided rewrites that replace the teacher template ~\citep{wu2025dap}. 
Beyond discrete trace editing, recent work has also explored shifting reasoning into continuous latent spaces~\citep{chen2025reasoninglanguagecomprehensivesurvey}, with particular attention to designing effective supervision signals for latent trajectories~\citep{chen2026makeseffectivesupervisionlatent}.
Orthogonal to trace editing, \textit{length control} constrains or predicts reasoning length, from lightweight inference-time constraints ~\citep{xu2025chain-of-draft, Ding:2024cotshortcut} to budget-aware methods that distill budget-aligned traces as supervision to internalize efficiency ~\citep{han2025tokenbudget}. 
Our work systematically characterizes this compression spectrum to map the trade-offs between structural fidelity and distillation performance.

\section{Experimental Setup}
\label{sec:Experiment}

\subsection{Problem Formulation}
We represent CoT compression as:
\begin{equation}
\label{eq:rewrite}
\tilde{r_i} = \mathcal{C}(r_i;\rho,\phi),
\end{equation}
which maps a teacher trace $r$ to a compressed version $\tilde{r}$ for fine-tuning. Here, $\rho$ denotes the compression level and $\phi$ captures methodological choices. 

\paragraph{Design Dimensions} We organize methods and experiments along three dimensions:

(i) \textbf{Importance criterion} $\phi_{\text{importance}}$:
Determines \emph{what} content is removed via a score-threshold-prune pipeline: each unit (token or step) is assigned an importance score, and those below a threshold derived from the target compression ratio are pruned. We instantiate three importance scores: LLMLingua2 (learned saliency), Entropy (uncertainty), and NLL (surprise).

(ii) \textbf{Restructuring level} $\phi_{\text{restruct}}$: Characterizes \emph{how} the trace is restructured, along a spectrum from surface trimming to template reconstruction. We use Gemini 2.5 Flash~\citep{comanici2025gemini25pushingfrontier} as the rewriter to span four levels: structural preservation (\textbf{L1: SP}), step editing (\textbf{L2: SE}), semantic fusion (\textbf{L3: SF}), and framework reconstruction (\textbf{L4: FR})~\footnote{Prompts and comparative cases are in Appendix~\ref{appendix:prompting_details} and Appendix~\ref{appendix:compress_cases}. We additionally verify using GPT-4o-mini~\citep{openai2024gpt4o} in Appendix~\ref{appendix:cross_rewriter}}.

(iii) \textbf{Compression Level $\rho$:} This factor governs the aggressiveness of supervision degradation. We manipulate a target ratio $\rho_{\text{target}}$ to observe the realized ratio $\rho_{\text{realized}}$--the actual trace length--which ultimately determines the end-to-end inference token ratio $\rho_{\text{inference}}$ as the deployment cost. This factor applies universally: as a threshold parameter for pruning or a length constraint for rewriting.

\paragraph{Objective and Evaluation.}
A student model $M_{\theta}$ is fine-tuned to minimize the standard negative log-likelihood on the compressed data $(x, \tilde{r}, y)$, derived from a specific teacher regime $\mathcal{R} \in \{\text{long}, \text{short}\}$.

{\sloppy
We evaluate the resulting impact surface defined by task performance ($\mathrm{Perf}$) and inference token cost ($\mathrm{Cost}$) across domains $d$:
\begin{equation}
\big(\mathrm{Perf}, \mathrm{Cost}\big) = f(\rho,\phi,\mathcal{R};d),
\end{equation}
where $\mathcal{R}$ denotes the training-trace regime (long/short) and $d$ the evaluation domain.}


\subsection{Models, Data, and Metrics}
\paragraph{Student Models.}
We employ LLaMA-3.1-8B-Instruct~\citep{grattafiori2024llama3} as the primary student model, with the Qwen2.5 series~\citep{yang2024qwen25} used to verify cross-model and scale generalization; full results are provided in Appendix~\ref{appendix:Result_equation}.

\paragraph{Training Corpora.} 
To analyze the impact of reasoning length, we construct two training sets: a Short-CoT regime sampled from Natural Reasoning~\citep{yuan2025naturalreasoningreasoningwild28m} (5k instances), and a Long-CoT regime from LIMO~\citep{ye2025limoreasoning}.

\paragraph{Evaluation Benchmarks.}
The evaluation spans five benchmarks in two domains:
(1) Math: GSM8K, MATH500 for \textit{short-CoT} regime and MATH500, AMC23, AIME24 for \textit{long-CoT} regime;(2) General: GPQA Diamond and MMLU-PRO for both regimes.

\paragraph{Evaluation Metrics.}
We report Accuracy for reasoning quality and the Inference Token Ratio ($\rho_{\text{inference}}$)—the ratio of generated tokens relative to the uncompressed baseline—for cost efficiency.

\paragraph{Implementation Details} We fine-tune all models via LoRA~\citep{lora}. Evaluation uses a standardized greedy-like decoding with temperature 0.7. Full hyperparameters, prompt templates for rewriting, and importance score definitions are provided in Appendix~\ref{appendix:training_details}.
\section{Effects of Importance Criterion}
\label{sec: Singal}

\begin{figure*}[t]
  \centering
  \includegraphics[width=1.0\linewidth]{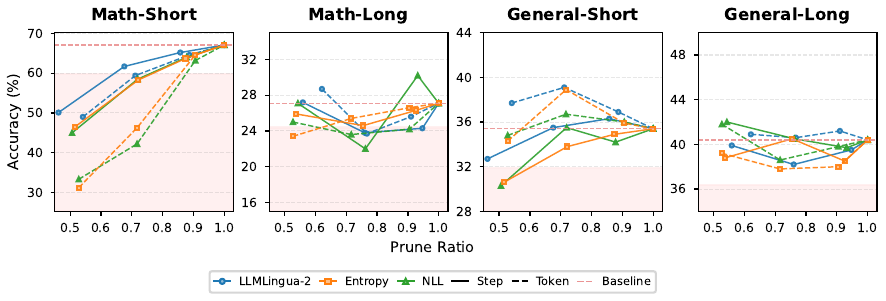}
  \caption{Accuracy of compressed students versus realized prune ratio for extraction-based pruning.}
  \label{fig:extraction_acc_ratio}
  \vspace{-12pt}
\end{figure*}

Whether CoT compression helps hinges on a simple premise: teacher traces must contain content that can be removed without changing the final performance, and the pruning mechanism must be able to target such content without discarding globally critical information.
Rather than estimating intrinsic information redundancy directly, we adopt an operational definition of compression tolerance: the capacity of a model to maintain downstream accuracy under increasing pruning rates.

\subsection{Empirical compression tolerance}

\begin{table}[t]
\centering
\small
\setlength{\tabcolsep}{24pt}
\begin{tabular}{lcc}
\toprule
Setting & Reflective steps (\%) \\
\midrule
Short-CoT & 0.57\%  \\
Long-CoT  & 30.2\%  \\
\bottomrule
\end{tabular}
\caption{
Content-level redundancy proxies for short- vs. long-CoT traces.
The percentage of reflective or self-correction steps per trace (e.g., ``wait'', ``let me check'').
}
\vspace{-12pt}
\label{tab:redundancy_proxies}
\end{table}


\paragraph{Long-CoT training exhibits higher empirical tolerance.}

As shown in Figure~\ref{fig:extraction_acc_ratio}, Long-CoT training yields a larger budget of removable tokens: student accuracy stays stable across a wider range of compressing ratios compared to Short-CoT. 
This is not because long traces contain a specific, easily-identifiable class of “redundant” steps, such as reflective statements(Table~\ref{tab:redundancy_proxies}) as suggested by prior work~\citep{jia2025epic}. In fact, in our controlled experiments (Appendix~\ref{app:ref_align}), pruning methods that deliberately target reflective statements perform no better than those that ignore them. Instead, redundancy in Long-CoT is diffuse: the critical reasoning skeleton is repeated and rephrased throughout the trace. Therefore, as long as that global skeleton survives, moderate compression is robust, irrespective of which specific elaborations are removed.

\paragraph{Domain-Dependent Sensitivity.} 
General benchmarks sustain accuracy under much stronger pruning than Math benchmarks. This is because General tasks typically accept diverse valid rationales and do not hinge on a specific intermediate state. Consequently, performance  slowly degrades even when the training signal is partially corrupted.

Math tasks, by contrast, require internally consistent intermediate quantities and precise constraint satisfaction. Small errors introduced by compression are amplified down the reasoning chain, leading to earlier and sharper accuracy drops under aggressive pruning~\footnote{Representative failure cases are in Appendix~\ref{appendix:failure_cases}}.

\begin{table}[t]
\centering
\small
\setlength{\tabcolsep}{4pt} 
\resizebox{\columnwidth}{!}{%
\begin{tabular}{lccc}
\toprule
Signal & Symbol / math & Stopwords & Other content \\
\midrule
Entropy      & 0.30 & 0.29 & 0.21 \\
NLL          & 0.30 & 0.28 & 0.21 \\
LLMLingua    & 0.15 & 0.54 & 0.11 \\
\bottomrule
\end{tabular}%
}
\caption{
The \textbf{realized} 0.3 target deleted ratio for \textsc{NaturalReasoning} under token-level pruning.
Compared with entropy/NLL, LLMLingua deletes fewer constraint-bearing tokens, shifting deletions toward stopwords.
}
\vspace{-12pt}
\label{tab:token_deletion_types}
\end{table}

\subsection{Importance Score Quality and Token Preservation}
\label{sec: Lingua_robust}

LLMLingua2 yields more stable pruning under high-risk compression.
At matched realized compression ratios, LLMLingua2-based pruning degrades more smoothly and suffers fewer sharp accuracy drops than entropy or NLL baselines. The gap is largest on Math benchmarks and under the strongest compression.

To understand why, we analyze which tokens each method deletes. We classify removed tokens into three categories: (i) math-like patterns (operators, numbers, logic markers), (ii) stopwords, and (iii) other content. Table~\ref{tab:token_deletion_types} shows that entropy and NLL delete a much larger fraction of math-like tokens than LLMLingua2, which concentrates its deletions on stopwords and other content. A full breakdown is in Appendix~\ref{app:token_analysis}.

Deleting math-like tokens is directly harmful. Failure cases~\footnote{Representative failure cases are in Appendix~\ref{appendix:failure_cases}} show that their removal correlates with catastrophic reasoning failures. Crucially, a controlled ablation~\footnote{Results are in Appendix~\ref{app:math_protection_ablation}} demonstrates that if we simply prevent entropy/NLL from deleting math tokens, their performance gap to LLMLingua2 largely disappears. This confirms that LLMLingua2’s advantage comes specifically from preserving symbolic anchors.

Thus, in high-risk extraction, the quality of the importance score dictates stability. Unlike probability-based heuristics that risk fracturing logical steps, LLMLingua2 functions as a symbol preserving filter, saving the symbolic anchors essential for valid derivation.

\begin{table}[t]
\centering
\small
\setlength{\tabcolsep}{6pt} 
\begin{tabular*}{\columnwidth}{@{\extracolsep{\fill}}lccc}
\toprule
Pair & 0.1 & 0.3 & 0.5 \\
\midrule
ENT--L2  & 0.75 & 0.53 & 0.36 \\
NLL--L2  & 0.76 & 0.54 & 0.38 \\
ENT--NLL & 0.92 & 0.85 & 0.78 \\
\bottomrule
\end{tabular*}
\caption{
Pairwise Jaccard overlap between preserved-step sets for different step-level signals under matched target ratios. (ENT: Entropy; L2: LLMLingua2)
}
\vspace{-12pt}
\label{tab:step-overlap}
\end{table}

\subsection{Granularity Modulates Importance Criterion Utility}
The utility of an importance criterion depends largely on the granularity of the deletion unit.
\paragraph{Step-level pruning: Convergence on the backbone.}
Despite their different scoring metrics, step-level strategies based on Entropy, NLL, and LLMLingua2 yield similar accuracy.
We analyze the pairwise Jaccard overlap of their pruning decisions, as reported in Table~\ref{tab:step-overlap}. 
The key insight is that the criteria agree strongly on which steps to preserve, where the overlap is 70–80\%, but disagree substantially on which steps to delete~\footnote{Details are in Appendix~\ref{app:step_overlap}}. This means that all criteria converge on the same critical reasoning backbone. The remaining steps are peripheral and interchangeable; removing one versus another has little effect. Hence, step-level pruning is robust to the choice of importance criterion.

\paragraph{Token-level pruning: importance score quality matters.}
In stark contrast, token-level pruning is highly sensitive to importance score quality. 
Under strong compression, LLMLingua2 maintains a smooth accuracy curve, while Entropy and NLL baselines collapse, especially on Math. This divergence comes down to whether the criterion preserves local structural integrity. 
As shown in Section~\ref{sec: Lingua_robust}, LLMLingua2 preferentially retains symbolic anchors and syntactic structures. Probability-based heuristics often indiscriminately delete key operators and numbers alongside stopwords. This risks fragmenting the internal logic of a reasoning step rather than trimming verbose phrasing, a pattern we confirm via coherence analysis~\footnote{Results and analysis details are in Appendix~\ref{app:coherence}}.

\begin{figure*}[t]
  \centering
  \includegraphics[width=1\linewidth]{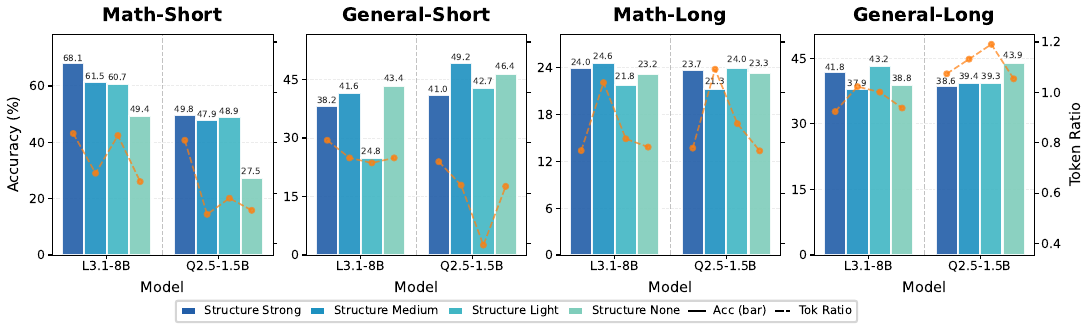}
  \caption{Accuracy and compression ratio across restructuring levels. Math exhibits a strict monotonic dependency on restructuring level (monotonic decay from SP to FR), whereas General domains show a restructuring benefit, where aggressive rewriting (SF/FR) matches or exceeds structure-preserving baselines.}
  \label{fig:structure_model_lineplot}
  \vspace{-12pt}
\end{figure*}

\mybox{{\bf Conclusion}}{gray!40}{gray!10}{Importance criterion utility is governed by granularity. At step level, all criteria converge on a shared reasoning backbone, making criterion choice irrelevant. At token level, this robustness collapses: symbol-aware criteria, like LLMLingua2, thin redundancy without shattering the logical core, while probability-based heuristics risk removing constraint-bearing tokens.}

\section{Effect of Restructuring Level}
\label{sec:Structure}


CoT compression changes not only length but also the organization of the reasoning chain.
We use the \textbf{restructuring level} to describe how much a compressed CoT preserves the teacher’s step segmentation, local ordering, and high-level solution plan.
We define a qualitative spectrum — from minimal to maximal perturbation — to characterize how each rewriting method restructures the reasoning path, while keeping compression ratio comparable.
Each level below represents an existing class of rewriting strategies, with representative work cited in line with prompts and comparative cases in Appendix~\ref{appendix:prompting_details} and Appendix~\ref{appendix:compress_cases}:

\begin{itemize}[leftmargin=*, itemsep=2pt]
\item \textbf{L1: Structural Preservation (SP):} The rewriter is constrained to only remove redundant words to target surface verbosity, strictly preserving the verbatim logical sequence~\citep{yu2025lsmixturesft}.
\item \textbf{L2: Structural Editing (SE):} The rewriter performs step-level operations like keeping, deleting or rephrasing to create a more efficient but structurally similar reasoning path~\citep{jiang2026drp}.
\item \textbf{L3: Semantic Fusion (SF):} The rewriter is encouraged to retain only ``key information'' to fuse multi-step sequences into novel statements~\citep{kang2025c3ot}.
\item \textbf{L4: Framework Replacement (FR):} The rewriter discards the original structure to regenerate reasoning via a predefined formal template~\citep{wu2025dap}.
\end{itemize}

\subsection{Robustness Across Restructuring Level}
As illustrated in Figure~\ref{fig:structure_model_lineplot}, a consistent pattern along this spectrum: \textit{across domains, lower restructuring levels are more robust as compression strengthens, whereas higher restructuring levels deteriorate earlier and show more domain-dependent variability.}

\paragraph{Robustness Degrades with Restructuring Level}
From SP to SE to SF as the restructuring level increases:
at light to moderate compression, SP/SE/SF typically reduce rationale length while staying near the original baseline, particularly on General tasks.
When compression becomes strong, however, SE and SF drop faster, while SP remains comparatively stable.
A plausible explanation is that SP/SE primarily shorten surface form while retaining the step skeleton, whereas SF already begins to fuse across steps; further compression then more easily weakens information that cannot be reconstructed from the remaining text.
At the highest restructuring level, framework replacement (FR) pushes this unstable effect further by rewriting the entire solution template, and we frequently observe higher variance across settings.

\subsection{Domain Sensitivity and Interaction}
\paragraph{Math Reasoning Demands Minimal Restructuring.} 
In LLM-based summarization, we observe a domain divergence. On Math, SF and SE degrade significantly faster than SP under strong compression, making SP the safest choice, as illustrated by representative failure cases in Appendix~\ref{appendix:failure_cases}. 
In Math domains, Accuracy declines monotonically as the restructuring level increases from SP to FR.
This fragility stems from the strict logical dependencies in mathematical reasoning: errors introduced by disrupting step boundaries through Semantic Fusion (L3) or by editing and rephrasing steps Structural Editing (L2) are amplified downstream. Conversely, General tasks often tolerate a wider range of valid rationales. Methods at higher restructuring levels remain competitive because these tasks are less sensitive to how intermediate steps are packaged, allowing for more aggressive abstraction.

\paragraph{FR: The High-Variance Extremity.} FR amplifies this domain split. On Math, it is among the weakest strategies; on General, it frequently reaches parity with, or even surpasses, structure-preserving methods under tight budgets. 
As shown in Figure~\ref{fig:structure_model_lineplot}, the Framework Replacement approach collapses on logic-heavy Math problems but remains competitive in General domains.
We attribute this to the rigidity of the solution space. Imposing a predefined framework (L4) on Math problems disrupts the precise logical flow of the original rationale, leading to derivation errors. In contrast, on General tasks, template-level rewriting can effectively reorganize and denoise the trace, yielding a more focused rationale.

\mybox{{\bf Conclusion}}{gray!40}{gray!10}{The effect of restructuring level inverts across domains: on Math, robustness degrades monotonically with restructuring level, leaving SP the only reliable choice. In General, this monotonicity breaks down: aggressive restructuring (SF/FR) can match or exceed structure-preserving baselines.}

\section{Effect of Compression Budget}
\label{sec:effect-ratio}

Building on the compression setup in Section~\ref{sec:Experiment}, we characterize compression dynamics via three metrics: the target ratio $\rho_{\text{target}}$ as the control knob, the empirical realized ratio $\rho_{\text{realized}}$ reflecting actual compression on CoT traces, and the inference token ratio $\rho_{\text{inference}}$ measuring deployment cost. Beyond fixed-$\rho_{\text{target}}$ constraints, we also evaluate a \emph{free ratio} mode, where the rewriter optimizes $\rho_{\text{realized}}$ naturally within a fixed structural template without a prescribed target.

\subsection{Stability and Performance Decay}
\begin{figure*}[t]
\centering
\includegraphics[width=1.0\textwidth]{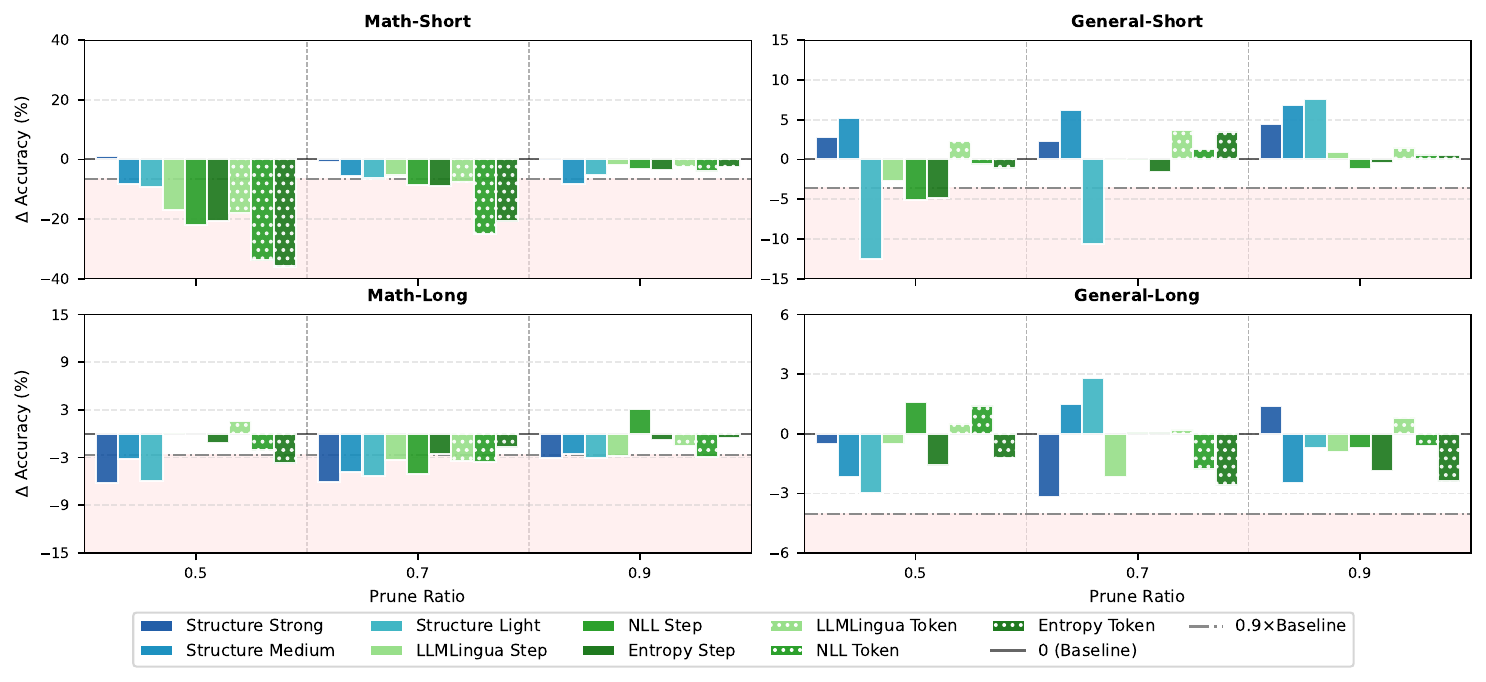}
\caption{Comparison of summarization- and extraction-based CoT compression across domains and training regimes. Each subplot reports accuracy under different prune ratios on Math/General $\times$ Short/Long settings. Bars show accuracy at prune ratios $0.5/0.7/0.9$, and the orange line shows the corresponding inference token ratio.}
\vspace{-12pt}
\label{fig:2*2_drop}
\end{figure*}

Our empirical analysis reveals that model accuracy remains remarkably resilient, staying within 90\% of the original baseline across a significant compression interval. However, as shown in Figure~\ref{fig:2*2_drop}, this stability is eventually followed by a sharp non-linear degradation. Once compression exceeds a critical threshold, performance collapses rapidly, suggesting a fundamental limit to how much information can be discarded before the underlying reasoning logic is compromised.

The compression resilience is modulated by two distinct factors.
Rather than a uniform decay, we observe that mathematical reasoning tasks are inherently more fragile than general-domain benchmarks, yet this sensitivity is significantly counteracted by long-CoT teachers, which systematically extend the viable compression range.

\begin{figure}[t]
\centering
\includegraphics[width=0.5\textwidth]{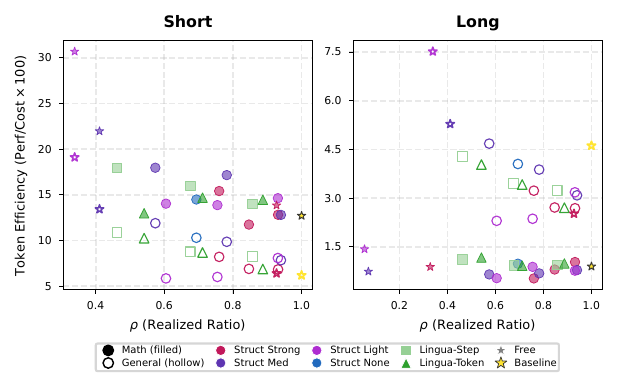}
\caption{Token efficiency ($\text{Perf}/\text{Cost}$) vs. Realized Ratio ($\rho_{\text{realized}}$), illustrating how structural fidelity defines length bands while the Free Ratio (stars) naturally identifies high-efficiency equilibrium points.}
\label{fig:token_efficiency}
\vspace{-12pt}
\end{figure}

\begin{figure}[t]
\centering
\includegraphics[width=0.5\textwidth]{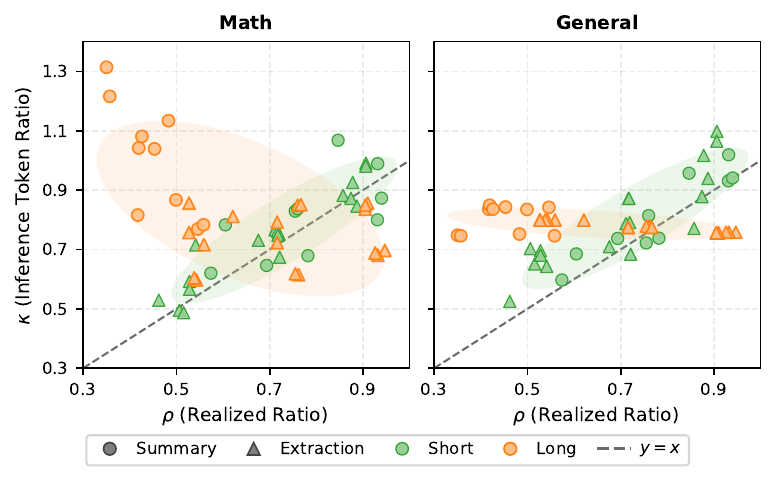}
\caption{Efficiency translation analysis. Short-CoT aligns with the $y=x$ diagonal while Long-CoT exhibits an efficiency gap ($\rho_{\text{inference}} > \rho_{\text{realized}}$).}
\label{fig:rho_kappa_scatter}
\vspace{-12pt}
\end{figure}

\subsection{Target vs. Realized Compression}

Given the observed resilience thresholds, we evaluate two distinct compression styles and employ a length-band analysis to characterize the resulting variability in token reduction.

\paragraph{Summarization as intent-driven rewriting.}
Unlike extractive methods, summarization modulates length indirectly via its rewriting objective.
We observe that realized lengths do not scale linearly with target ratios; instead, the teacher LLM, constrained by linguistic fluency and structural coherence, converges to method-specific characteristic lengths.
Thus, the target ratio signals compression intensity rather than a precise numerical constraint.

\paragraph{Structural constraints define the length band.} As shown in Figure~\ref{fig:token_efficiency}, the choice of structural fidelity dictates the achievable compression range (the band), while the target ratio only facilitates marginal adjustments within that band.

In High-fidelity methods, such as SP, strict structural constraints force the model to preserve the original reasoning skeleton, yielding a narrow, conservative band.
Looser prompts methods such as SF/SE, permissive constraints allowing cross-step fusion enable the model to fold multiple steps into single spans, shifting the band toward smaller ratios and widening its variance.
This suggests that stronger structural constraints yield stable but conservative results, while weaker constraints enable more aggressive but volatile compression.

\paragraph{Free Ratio as a Natural Equilibrium.} The free-ratio represents a method's intrinsic set-point absent of rigid constraints. Figure~\ref{fig:token_efficiency} confirms that this "natural" landing point often yields superior token efficiency, appearing as ``star'' points on the optimal frontier. In structure-preserving methods, this mechanism serves as a self-regularizer, ensuring the model settles at stable, moderate ratios. Although less protective for aggressive variants with naturally lower bands, the free ratio avoids the instability of forced compression. Thus, we propose using the free ratio as a primary configuration for ensuring consistent performance.

\subsection{The Gap Between Training and Inference}

We observe a positive correlation between the end-to-end inference token ratio and realized compression ratio. As shown in Figure~\ref{fig:rho_kappa_scatter}, this correlation is tight in the short-CoT regime but weakens under long-CoT regime.
Generally, $\rho_{\text{target}}$ serves as a viable proxy, but the efficiency translation varies significantly by regime.
Short-CoT exhibits a near-linear correlation ($\rho_{\text{inference}} \approx \rho_{\text{realized}}$) along the diagonal, indicating successful internalization of concise reasoning styles. 
In contrast, the Long-CoT regime reveals a significant efficiency gap where $\rho_{\text{inference}} > \rho_{\text{realized}}$. This divergence stems from a translation loss where verbose habits persist despite concise training supervision. Thus, for complex tasks, $\rho$ serves as an optimistic lower bound rather than a precise predictor of deployment cost.
For practical use, Appendix~\ref{app:decision_guide} consolidates the findings from Sections~4--6 into a condition-aware decision guide.


\mybox{{\bf Conclusion}}{gray!40}{gray!10}{
As compression budget decreases, accuracy degrades as a "plateau-then-drop", with resilience shaped by domain and source-CoT regime. More critically, efficiency translation is regime-conditional: Short-CoT compresses faithfully, while Long-CoT does not, as verbose habits persist despite concise supervision, making $\rho$ an optimistic lower bound for deployment cost.
}

\vspace{+24pt}
\section{Conclusion}
\label{sec:conclusion}

We have reframed CoT compression into three design dimensions: compression budget, importance criterion, and restructuring level. Each is governed by a distinct factor. Whether training-side compression translates to shorter inference depends on the source-CoT regime, with Long-CoT students retaining verbose habits despite concise supervision. Criterion utility is governed by granularity: step-level methods converge on a shared backbone, while token-level pruning demands symbol-aware criteria such as LLMLingua2 to preserve the logical core. Restructuring level acts in opposite directions across domains, denoising General tasks but degrading Math monotonically. Together, these results show that compression efficacy is conditioned on regime, granularity, and domain; no single strategy universally dominates. Effective deployment requires aligning each design choice with its specific task context and inference constraints.

\section*{Limitations}
\label{subsec:limitation} 
Our experiments adopt LoRA-based fine-tuning as the training 
objective, as our goal is to understand how different compression strategies alter the supervision signal itself rather than to compensate for supervision degradation through stronger distillation objectives. Full-parameter fine-tuning and more aggressive objectives remain important directions for future work. Furthermore, our Math/General partition is a pragmatic dichotomy that isolates two prototypical classes of structural sensitivity rather than an exhaustive taxonomy of reasoning task types. More specific task types, such as code generation, deep-search reasoning, and multi-hop problems, may exhibit hybrid sensitivity combining symbolic dependencies with more flexible solution spaces, suggesting that the General category could be further decomposed along structural properties. We leave such a finer-grained taxonomy as a concrete direction for future work.

\section*{Ethics Statement}
\label{subsec:ethics} 
The datasets utilized in this study are publicly available benchmarks. We acknowledge, however, that public web-sourced data may inherently contain socially harmful or toxic language, despite our inability to guarantee its complete absence. Our experiments rely on the Google Gemini API; at no point did we access or attempt to access the proprietary training data or underlying system weights of the models. We confirm that the datasets used do not contain personally identifiable information (PII), ensuring user privacy. All datasets and models were used in strict accordance with their permissive licenses and intended usage guidelines. We believe this work adheres to the ACL Code of Ethics. Finally, we employed GPT-5\footnote{\url{https://chatgpt.com/}}to correct grammatical errors in this paper.

\section*{Acknowledgments}
\label{subsec:acknowledgments} 
We thank the Center for Advanced Research Computing (CARC) at the University of Southern California for providing the computational resources used in this work.

\bibliography{custom}

@inproceedings{yuan2025naturalreasoningreasoningwild28m,
  title     = {{NaturalReasoning}: Reasoning in the Wild with {2.8M} Challenging Questions},
  author    = {Yuan, Weizhe and Yu, Jane and Jiang, Song and Padthe, Karthik
               and Li, Yang and Wang, Dong and Kulikov, Ilia and Cho, Kyunghyun
               and Tian, Yuandong and Weston, Jason E. and Li, Xian},
  booktitle = {Advances in Neural Information Processing Systems},
  volume    = {38},
  pages     = {74615--74646},
  year      = {2025},
  doi       = {10.52202/085713-2246},
  url       = {https://papers.neurips.cc/paper_files/paper/2025/hash/60b22138917015fcba7165670e6f453c-Abstract-Datasets_and_Benchmarks_Track.html}
}

@misc{zhen2025reconsideringoverthinking,
  title         = {Reconsidering Overthinking: Penalizing Internal
                   and External Redundancy in {CoT} Reasoning},
  author        = {Taihang Zhen and Jialiang Hong and Kai Chen and
                   Guang Yang and Junlan Feng and Wenpeng Zhu and
                   Jing Huo and Yang Gao and Depeng Wang and Haitao Wan
                   and Xi Yang and Fanyu Meng and Yuyao Zhang and
                   Ji Qi and Xiangyu Zhou},
  year          = {2025},
  eprint        = {2508.02178},
  archivePrefix = {arXiv},
  primaryClass  = {cs.AI},
  doi           = {10.48550/arXiv.2508.02178},
  url           = {https://arxiv.org/abs/2508.02178}
}

@inproceedings{ye2025limoreasoning,
  title     = {{LIMO}: Less is More for Reasoning},
  author    = {Yixin Ye and Zhen Huang and Yang Xiao and
               Ethan Chern and Shijie Xia and Pengfei Liu},
  booktitle = {Conference on Language Modeling},
  year      = {2025},
  url       = {https://openreview.net/forum?id=T2TZ0RY4Zk}
}

@misc{grattafiori2024llama3,
  title         = {The {Llama 3} Herd of Models},
  author        = {Grattafiori, Aaron and Dubey, Abhimanyu and Jauhri, Abhinav and Pandey, Abhinav and Kadian, Abhishek and Al-Dahle, Ahmad and others},
  year          = {2024},
  month         = jul,
  eprint        = {2407.21783},
  archivePrefix = {arXiv},
  primaryClass  = {cs.AI},
  doi           = {10.48550/arXiv.2407.21783},
  url           = {https://arxiv.org/abs/2407.21783}
}

@misc{comanici2025gemini25pushingfrontier,
      title={Gemini 2.5: Pushing the Frontier with Advanced Reasoning, Multimodality, Long Context, and Next Generation Agentic Capabilities}, 
      author={{Google DeepMind Gemini Team}},
      year={2025},
      eprint={2507.06261},
      archivePrefix={arXiv},
      primaryClass={cs.CL},
      url={https://arxiv.org/abs/2507.06261}, 
}

@misc{wang2025r1compress,
  title         = {{R1-Compress}: Long Chain-of-Thought Compression via Chunk Compression and Search},
  author        = {Wang, Yibo and Shen, Li and Yao, Huanjin and Huang, Tiansheng
                   and Liu, Rui and Tan, Naiqiang and Huang, Jiaxing
                   and Zhang, Kai and Tao, Dacheng},
  year          = {2025},
  month         = may,
  eprint        = {2505.16838},
  archivePrefix = {arXiv},
  primaryClass  = {cs.CL},
  doi           = {10.48550/arXiv.2505.16838},
  url           = {https://arxiv.org/abs/2505.16838}
}

@article{kang2025c3ot,
  title   = {{C3oT}: Generating Shorter Chain-of-Thought Without Compromising Effectiveness},
  author  = {Kang, Yu and Sun, Xianghui and Chen, Liangyu and Zou, Wei},
  journal = {Proceedings of the AAAI Conference on Artificial Intelligence},
  volume  = {39},
  number  = {23},
  pages   = {24312--24320},
  year    = {2025},
  doi     = {10.1609/aaai.v39i23.34608},
  url     = {https://doi.org/10.1609/aaai.v39i23.34608}
}

@inproceedings{han2025tokenbudget,
  title     = {Token-Budget-Aware {LLM} Reasoning},
  author    = {Han, Tingxu and Wang, Zhenting and Fang, Chunrong and Zhao, Shiyu and Ma, Shiqing and Chen, Zhenyu},
  booktitle = {Findings of the Association for Computational Linguistics: ACL 2025},
  month     = jul,
  year      = {2025},
  address   = {Vienna, Austria},
  publisher = {Association for Computational Linguistics},
  pages     = {24842--24855},
  doi       = {10.18653/v1/2025.findings-acl.1274},
  url       = {https://aclanthology.org/2025.findings-acl.1274/}
}

@misc{wu2025dap,
  title         = {Concise Reasoning, Big Gains: Pruning Long Reasoning Trace with Difficulty-Aware Prompting},
  author        = {Wu, Yifan and Shi, Jingze and Wu, Bingheng and Zhang, Jiayi and Lin, Xiaotian and Tang, Nan and Luo, Yuyu},
  year          = {2025},
  month         = may,
  eprint        = {2505.19716},
  archivePrefix = {arXiv},
  primaryClass  = {cs.AI},
  url           = {https://arxiv.org/abs/2505.19716}
}

@inproceedings{jiang2026drp,
  title     = {{DRP}: Distilled Reasoning Pruning with Mathematical Skill-aware Step Decomposition for Efficient Large Reasoning Models},
  author    = {Jiang, Yuxuan and Li, Dawei and Ferraro, Francis},
  booktitle = {Findings of the Association for Computational Linguistics: ACL 2026},
  month     = jul,
  year      = {2026},
  address   = {San Diego, California, United States},
  publisher = {Association for Computational Linguistics},
  pages     = {4020--4039},
  doi       = {10.18653/v1/2026.findings-acl.196},
  url       = {https://aclanthology.org/2026.findings-acl.196/}
}

@misc{yu2025lsmixturesft,
  title         = {Long-Short Chain-of-Thought Mixture Supervised Fine-Tuning Eliciting Efficient Reasoning in Large Language Models},
  author        = {Yu, Bin and Yuan, Hang and Li, Haotian and Xu, Xueyin and Wei, Yuliang and Wang, Bailing and Qi, Weizhen and Chen, Kai},
  year          = {2025},
  month         = may,
  eprint        = {2505.03469},
  archivePrefix = {arXiv},
  primaryClass  = {cs.CL},
  url           = {https://arxiv.org/abs/2505.03469}
}

@inproceedings{xu2025astarthought,
  title     = {{A\*-Thought}: Efficient Reasoning via Bidirectional Compression for Low-Resource Settings},
  author    = {Xu, Xiaoang and Wang, Shuo and Han, Xu and Liu, Zhenghao and Wu, Huijia and Li, Peipei and Liu, Zhiyuan and Sun, Maosong and He, Zhaofeng},
  booktitle = {Advances in Neural Information Processing Systems},
  volume    = {38},
  year      = {2025},
  doi       = {10.52202/085713-3594},
  url       = {https://proceedings.neurips.cc/paper_files/paper/2025/hash/9ac5238b5c033fc49c2932a50df2bd47-Abstract-Conference.html}
}

@misc{zhao2025pruneonlogic,
  title         = {Can Pruning Improve Reasoning? Revisiting Long-{CoT} Compression with Capability in Mind for Better Reasoning},
  author        = {Zhao, Shangziqi and Yuan, Jiahao and Wu, Jinyang and Wang, Zhenglin and Yang, Guisong and Naseem, Usman},
  year          = {2025},
  month         = may,
  eprint        = {2505.14582},
  archivePrefix = {arXiv},
  primaryClass  = {cs.CL},
  doi           = {10.48550/arXiv.2505.14582},
  url           = {https://arxiv.org/abs/2505.14582}
}

@inproceedings{cui2025stepwiseperplexity,
  title     = {Stepwise Perplexity-Guided Refinement for Efficient Chain-of-Thought Reasoning in Large Language Models},
  author    = {Cui, Yingqian and He, Pengfei and Zeng, Jingying and Liu, Hui and Tang, Xianfeng and Dai, Zhenwei and Han, Yan and Luo, Chen and Huang, Jing and Li, Zhen and Wang, Suhang and Xing, Yue and Tang, Jiliang and He, Qi},
  booktitle = {Findings of the Association for Computational Linguistics: ACL 2025},
  month     = jul,
  year      = {2025},
  address   = {Vienna, Austria},
  publisher = {Association for Computational Linguistics},
  pages     = {18581--18597},
  doi       = {10.18653/v1/2025.findings-acl.956},
  url       = {https://aclanthology.org/2025.findings-acl.956/}
}

@inproceedings{li2026stepentropy,
  title     = {Making Slow Thinking Faster: Compressing {LLM} Chain-of-Thought via Step Entropy},
  author    = {Li, Zeju and Zhong, Jianyuan and Zheng, Ziyang and Wen, Xiangyu and Xu, Zhijian and Cheng, Yingying and Zhang, Fan and Xu, Qiang},
  booktitle = {The Fourteenth International Conference on Learning Representations},
  year      = {2026},
  url       = {https://openreview.net/forum?id=cGLqQfS5wH}
}

@inproceedings{xia2025tokenskip,
  title     = {{T}oken{S}kip: Controllable Chain-of-Thought
               Compression in {LLM}s},
  author    = {Xia, Heming and Leong, Chak Tou and Wang, Wenjie and
               Li, Yongqi and Li, Wenjie},
  booktitle = {Proceedings of the 2025 Conference on Empirical
               Methods in Natural Language Processing},
  month     = nov,
  year      = {2025},
  address   = {Suzhou, China},
  publisher = {Association for Computational Linguistics},
  pages     = {3351--3363},
  doi       = {10.18653/v1/2025.emnlp-main.165},
  url       = {https://aclanthology.org/2025.emnlp-main.165/}
}

@misc{jia2025epic,
  title         = {{EPiC}: Towards Lossless Speedup for Reasoning Training through Edge-Preserving CoT Condensation},
  author        = {Jia, Jinghan and Reisizadeh, Hadi and Fan, Chongyu and Baracaldo, Nathalie and Hong, Mingyi and Liu, Sijia},
  year          = {2025},
  month         = jun,
  eprint        = {2506.04205},
  archivePrefix = {arXiv},
  primaryClass  = {cs.LG},
  url           = {https://arxiv.org/abs/2506.04205}
}

@inproceedings{wang2024mmlupro,
  title     = {{MMLU-Pro}: A More Robust and Challenging Multi-Task Language Understanding Benchmark},
  author    = {Wang, Yubo and Ma, Xueguang and Zhang, Ge and Ni, Yuansheng
               and Chandra, Abhranil and Guo, Shiguang and Ren, Weiming
               and Arulraj, Aaran and He, Xuan and Jiang, Ziyan and Li, Tianle
               and Ku, Max and Wang, Kai and Zhuang, Alex and Fan, Rongqi
               and Yue, Xiang and Chen, Wenhu},
  booktitle = {Advances in Neural Information Processing Systems},
  volume    = {37},
  pages     = {95266--95290},
  year      = {2024},
  doi       = {10.52202/079017-3018},
  url       = {https://papers.nips.cc/paper_files/paper/2024/hash/ad236edc564f3e3156e1b2feafb99a24-Abstract-Datasets_and_Benchmarks_Track.html}
}

@inproceedings{rein2024gpqa,
  title     = {GPQA: A Graduate-Level Google-Proof Q\&A Benchmark},
  author    = {Rein, David and Hou, Betty Li and Stickland, Asa Cooper and Petty, Jackson and Pang, Richard Yuanzhe and Dirani, Julien and Michael, Julian and Bowman, Samuel R.},
  booktitle = {First Conference on Language Modeling},
  year      = {2024},
  url       = {https://openreview.net/forum?id=Ti67584b98}
}

@misc{aops2025aimeproblems,
  author = {{Art of Problem Solving}},
  title  = {{AIME} Problems and Solutions},
  year   = {n.d.},
  url    = {https://artofproblemsolving.com/wiki/index.php/AIME_Problems_and_Solutions},
  note   = {Accessed: 2025-12-23}
}

@misc{xu2025chain-of-draft,
  title         = {Chain of Draft: Thinking Faster by Writing Less},
  author        = {Xu, Silei and Xie, Wenhao and Zhao, Lingxiao and He, Pengcheng},
  year          = {2025},
  month         = feb,
  eprint        = {2502.18600},
  archivePrefix = {arXiv},
  primaryClass  = {cs.CL},
  doi           = {10.48550/arXiv.2502.18600},
  url           = {https://arxiv.org/abs/2502.18600}
}

@article{chen-etal-2025-towards-reasoning-era,
  title   = {Towards Reasoning Era: A Survey of Long Chain-of-Thought for Reasoning Large Language Models},
  author  = {Chen, Qiguang and Qin, Libo and Liu, Jinhao and Peng, Dengyun
             and Guan, Jiannan and Wang, Peng and Hu, Mengkang and Zhou, Yuhang
             and Gao, Te and Che, Wanxiang},
  journal = {Science China Information Sciences},
  volume  = {69},
  number  = {6},
  pages   = {161101},
  year    = {2026},
  month   = may,
  doi     = {10.1007/s11432-025-4665-8},
  url     = {https://doi.org/10.1007/s11432-025-4665-8}
}

@misc{deepseekai2025deepseekr1incentivizingreasoningcapability,
      title={DeepSeek-R1: Incentivizing Reasoning Capability in LLMs via Reinforcement Learning}, 
      author={{DeepSeek-AI} and Daya Guo and others},
      year={2025},
      eprint={2501.12948},
      archivePrefix={arXiv},
      primaryClass={cs.CL},
      url={https://arxiv.org/abs/2501.12948}, 
}

@misc{huang2024o1replicationjourney,
      title={O1 Replication Journey -- Part 2: Surpassing O1-preview through Simple Distillation, Big Progress or Bitter Lesson?}, 
      author={Zhen Huang and Haoyang Zou and Xuefeng Li and Yixiu Liu and Yuxiang Zheng and Ethan Chern and Shijie Xia and Yiwei Qin and Weizhe Yuan and Pengfei Liu},
      year={2024},
      eprint={2411.16489},
      archivePrefix={arXiv},
      primaryClass={cs.CL},
      url={https://arxiv.org/abs/2411.16489}, 
}

@article{sui2025stopoverthinkingsurveyefficient,
  title   = {Stop Overthinking: A Survey on Efficient Reasoning for Large Language Models},
  author  = {Sui, Yang and Chuang, Yu-Neng and Wang, Guanchu and Zhang, Jiamu and Zhang, Tianyi and Yuan, Jiayi and Liu, Hongyi and Wen, Andrew and Zhong, Shaochen and Zou, Na and Chen, Hanjie and Hu, Xia},
  journal = {Transactions on Machine Learning Research},
  year    = {2025},
  issn    = {2835-8856},
  url     = {https://openreview.net/forum?id=HvoG8SxggZ}
}

@misc{wang2025harnessingreasoningeconomysurvey,
      title={Harnessing the Reasoning Economy: A Survey of Efficient Reasoning for Large Language Models}, 
      author={Rui Wang and Hongru Wang and Boyang Xue and Jianhui Pang and Shudong Liu and Yi Chen and Jiahao Qiu and Derek Fai Wong and Heng Ji and Kam-Fai Wong},
      year={2025},
      eprint={2503.24377},
      archivePrefix={arXiv},
      primaryClass={cs.CL},
      url={https://arxiv.org/abs/2503.24377}, 
}

@inproceedings{chen2025think23overthinkingo1like,
  title     = {Do {NOT} Think That Much for 2+3=? On the
               Overthinking of Long Reasoning Models},
  author    = {Xingyu Chen and Jiahao Xu and Tian Liang and
               Zhiwei He and Jianhui Pang and Dian Yu and Linfeng Song
               and Qiuzhi Liu and Mengfei Zhou and Zhuosheng Zhang and
               Rui Wang and Zhaopeng Tu and Haitao Mi and Dong Yu},
  booktitle = {Proceedings of the 42nd International Conference on
               Machine Learning},
  pages     = {9487--9499},
  year      = {2025},
  volume    = {267},
  series    = {Proceedings of Machine Learning Research},
  publisher = {PMLR},
  url       = {https://proceedings.mlr.press/v267/chen25bx.html}
}

@inproceedings{wang2025thoughtsplaceunderthinking,
  title     = {Thoughts Are All Over the Place: On the
               Underthinking of Long Reasoning Models},
  author    = {Yue Wang and Qiuzhi Liu and Jiahao Xu and Tian Liang
               and Xingyu Chen and Zhiwei He and Linfeng Song and
               Dian Yu and Juntao Li and Zhuosheng Zhang and Rui Wang
               and Zhaopeng Tu and Haitao Mi and Dong Yu},
  booktitle = {Advances in Neural Information Processing Systems},
  volume    = {38},
  year      = {2025},
  doi       = {10.52202/085713-1027},
  url       = {https://proceedings.neurips.cc/paper_files/paper/2025/hash/2bfc4acb2e5c4af1576d14b0df9a664c-Abstract-Conference.html}
}

@misc{openai2024_learning_reason_llms,
  author = {{OpenAI}},
  title  = {Learning to Reason with {LLMs}},
  year   = {2024},
  month  = sep,
  day    = {12},
  url    = {https://openai.com/index/learning-to-reason-with-llms/},
  note   = {Accessed: 2025-12-30}
}

@misc{zhu2025conciseadaptivethinkinglarge,
      title={Towards Concise and Adaptive Thinking in Large Reasoning Models: A Survey}, 
      author={Jason Zhu and Hongyu Li},
      year={2025},
      eprint={2507.09662},
      archivePrefix={arXiv},
      primaryClass={cs.AI},
      url={https://arxiv.org/abs/2507.09662}, 
}

@inproceedings{chen-etal-2025-unveiling-key,
    title = "Unveiling the Key Factors for Distilling Chain-of-Thought Reasoning",
    author = "Chen, Xinghao  and
      Sun, Zhijing  and
      Wenjin, Guo  and
      Zhang, Miaoran  and
      Chen, Yanjun  and
      Sun, Yirong  and
      Su, Hui  and
      Pan, Yijie  and
      Klakow, Dietrich  and
      Li, Wenjie  and
      Shen, Xiaoyu",
    editor = "Che, Wanxiang  and
      Nabende, Joyce  and
      Shutova, Ekaterina  and
      Pilehvar, Mohammad Taher",
    booktitle = "Findings of the Association for Computational Linguistics: ACL 2025",
    month = jul,
    year = "2025",
    address = "Vienna, Austria",
    publisher = "Association for Computational Linguistics",
    url = "https://aclanthology.org/2025.findings-acl.782/",
    doi = "10.18653/v1/2025.findings-acl.782",
    pages = "15094--15119",
    ISBN = "979-8-89176-256-5"
}

@inproceedings{dao2024flashattention2,
  title     = {{FlashAttention-2}: Faster Attention with Better Parallelism and Work Partitioning},
  author    = {Dao, Tri},
  booktitle = {International Conference on Learning Representations},
  year      = {2024},
  url       = {https://openreview.net/forum?id=mZn2Xyh9Ec}
}

@inproceedings{rajbhandari2020zero,
  author    = {Samyam Rajbhandari and Jeff Rasley and Olatunji Ruwase and Yuxiong He},
  title     = {ZeRO: Memory Optimizations Toward Training Trillion Parameter Models},
  booktitle = {Proceedings of the International Conference for High Performance Computing, Networking, Storage and Analysis (SC '20)},
  year      = {2020},
  publisher = {IEEE/ACM},
  articleno = {20},
  numpages  = {16},
  doi       = {10.1109/SC41405.2020.00024},
  url       = {https://dl.acm.org/doi/10.5555/3433701.3433727}
}

@inproceedings{bird-loper-2004-nltk,
  title = "{NLTK}: The Natural Language Toolkit",
  author = "Bird, Steven  and
    Loper, Edward",
  booktitle = "Proceedings of the {ACL} Interactive Poster and Demonstration Sessions",
  month = jul,
  year = "2004",
  address = "Barcelona, Spain",
  publisher = "Association for Computational Linguistics",
  url = "https://aclanthology.org/P04-3031/",
  pages = "214--217"
}

@inproceedings{pan:2024llmlingua2,
    title = "{LLML}ingua-2: Data Distillation for Efficient and Faithful Task-Agnostic Prompt Compression",
    author = {Pan, Zhuoshi  and
      Wu, Qianhui  and
      Jiang, Huiqiang  and
      Xia, Menglin  and
      Luo, Xufang  and
      Zhang, Jue  and
      Lin, Qingwei  and
      R{\"u}hle, Victor  and
      Yang, Yuqing  and
      Lin, Chin-Yew  and
      Zhao, H. Vicky  and
      Qiu, Lili  and
      Zhang, Dongmei},
    editor = "Ku, Lun-Wei  and
      Martins, Andre  and
      Srikumar, Vivek",
    booktitle = "Findings of the Association for Computational Linguistics: ACL 2024",
    month = aug,
    year = "2024",
    address = "Bangkok, Thailand",
    publisher = "Association for Computational Linguistics",
    url = "https://aclanthology.org/2024.findings-acl.57/",
    doi = "10.18653/v1/2024.findings-acl.57",
    pages = "963--981",
}

@article{Ding:2024cotshortcut,
  author       = {Mengru Ding and
                  Hanmeng Liu and
                  Zhizhang Fu and
                  Jian Song and
                  Wenbo Xie and
                  Yue Zhang},
  title        = {Break the Chain: Large Language Models Can be Shortcut Reasoners},
  journal      = {CoRR},
  volume       = {abs/2406.06580},
  year         = {2024},
  url          = {https://doi.org/10.48550/arXiv.2406.06580},
  doi          = {10.48550/ARXIV.2406.06580},
  eprinttype    = {arXiv},
  eprint       = {2406.06580},
  bibsource    = {dblp computer science bibliography, https://dblp.org}
}

@article{yang2024qwen25,
  title   = {{Qwen2.5} Technical Report},
  author  = {Yang, An and Yang, Baosong and Zhang, Beichen and Hui, Binyuan
             and Zheng, Bo and Yu, Bowen and Li, Chengyuan and Liu, Dayiheng
             and Huang, Fei and Wei, Haoran and Lin, Huan and Yang, Jian
             and Tu, Jianhong and Zhang, Jianwei and Yang, Jianxin
             and Yang, Jiaxi and Zhou, Jingren and Lin, Junyang and Dang, Kai
             and Lu, Keming and Bao, Keqin and Yang, Kexin and Yu, Le
             and Li, Mei and Xue, Mingfeng and Zhang, Pei and Zhu, Qin
             and Men, Rui and Lin, Runji and Li, Tianhao and Tang, Tianyi
             and Xia, Tingyu and Ren, Xingzhang and Ren, Xuancheng
             and Fan, Yang and Su, Yang and Zhang, Yichang and Wan, Yu
             and Liu, Yuqiong and Cui, Zeyu and Zhang, Zhenru and Qiu, Zihan},
  journal = {CoRR},
  volume  = {abs/2412.15115},
  year    = {2024},
  doi     = {10.48550/arXiv.2412.15115},
  url     = {https://arxiv.org/abs/2412.15115}
}

@inproceedings{lora,
  author       = {Edward J. Hu and
                  Yelong Shen and
                  Phillip Wallis and
                  Zeyuan Allen{-}Zhu and
                  Yuanzhi Li and
                  Shean Wang and
                  Lu Wang and
                  Weizhu Chen},
  title        = {LoRA: Low-Rank Adaptation of Large Language Models},
  booktitle    = {The Tenth International Conference on Learning Representations, {ICLR}
                  2022, Virtual Event, April 25-29, 2022},
  publisher    = {OpenReview.net},
  year         = {2022},
  url          = {https://openreview.net/forum?id=nZeVKeeFYf9},
  bibsource    = {dblp computer science bibliography, https://dblp.org}
}

@inproceedings{llamafactory,
    title = "{L}lama{F}actory: Unified Efficient Fine-Tuning of 100+ Language Models",
    author = "Zheng, Yaowei  and
      Zhang, Richong  and
      Zhang, Junhao  and
      Ye, Yanhan  and
      Luo, Zheyan",
    editor = "Cao, Yixin  and
      Feng, Yang  and
      Xiong, Deyi",
    booktitle = "Proceedings of the 62nd Annual Meeting of the Association for Computational Linguistics (Volume 3: System Demonstrations)",
    month = aug,
    year = "2024",
    address = "Bangkok, Thailand",
    publisher = "Association for Computational Linguistics",
    url = "https://aclanthology.org/2024.acl-demos.38/",
    doi = "10.18653/v1/2024.acl-demos.38",
    pages = "400--410"
}

@article{Cobbe:2021gsm8k,
  author       = {Karl Cobbe and
                  Vineet Kosaraju and
                  Mohammad Bavarian and
                  Mark Chen and
                  Heewoo Jun and
                  Lukasz Kaiser and
                  Matthias Plappert and
                  Jerry Tworek and
                  Jacob Hilton and
                  Reiichiro Nakano and
                  Christopher Hesse and
                  John Schulman},
  title        = {Training Verifiers to Solve Math Word Problems},
  journal      = {CoRR},
  volume       = {abs/2110.14168},
  year         = {2021},
  url          = {https://arxiv.org/abs/2110.14168},
  eprinttype    = {arXiv},
  eprint       = {2110.14168},
  bibsource    = {dblp computer science bibliography, https://dblp.org}
}

@inproceedings{math,
    title={Measuring Mathematical Problem Solving With the {MATH} Dataset},
    author={Dan Hendrycks and Collin Burns and Saurav Kadavath and Akul Arora and Steven Basart and Eric Tang and Dawn Song and Jacob Steinhardt},
    booktitle={Thirty-fifth Conference on Neural Information Processing Systems Datasets and Benchmarks Track (Round 2)},
    year={2021},
    url={https://openreview.net/forum?id=7Bywt2mQsCe}
}

@misc{maa_amc_2025,
  author = {{Mathematical Association of America}},
  title  = {{American Mathematics Competitions (AMC)}},
  year   = {2025},
  url    = {https://maa.org/student-programs/amc/},
  note   = {Accessed: 2026-01-06}
}

@misc{openai2024gpt4o,
  title         = {{GPT-4o} System Card},
  author        = {{OpenAI}},
  year          = {2024},
  month         = oct,
  eprint        = {2410.21276},
  archivePrefix = {arXiv},
  primaryClass  = {cs.CL},
  doi           = {10.48550/arXiv.2410.21276},
  url           = {https://arxiv.org/abs/2410.21276}
}

@inproceedings{chen2026makeseffectivesupervisionlatent,
  title     = {What Makes Effective Supervision in Latent Chain-of-Thought: An Information-Theoretic Analysis},
  author    = {Chen, Xinghao and Leong, Chak Tou and Guo, Wenjin and Wang, Jian and Li, Wenjie and Shen, Xiaoyu},
  booktitle = {Forty-third International Conference on Machine Learning},
  year      = {2026},
  url       = {https://openreview.net/forum?id=x21YkF1UfR}
}

@misc{chen2025reasoninglanguagecomprehensivesurvey,
      title={Reasoning Beyond Language: A Comprehensive Survey on Latent Chain-of-Thought Reasoning}, 
      author={Xinghao Chen and Anhao Zhao and Heming Xia and Xuan Lu and Hanlin Wang and Yanjun Chen and Wei Zhang and Jian Wang and Wenjie Li and Xiaoyu Shen},
      year={2025},
      eprint={2505.16782},
      archivePrefix={arXiv},
      primaryClass={cs.CL},
      url={https://arxiv.org/abs/2505.16782}, 
}

@misc{ding2026llmslrmsrethinkingpruning,
      title={From LLMs to LRMs: Rethinking Pruning for Reasoning-Centric Models}, 
      author={Longwei Ding and Anhao Zhao and Fanghua Ye and Ziyang Chen and Xiaoyu Shen},
      year={2026},
      eprint={2601.18091},
      archivePrefix={arXiv},
      primaryClass={cs.LG},
      url={https://arxiv.org/abs/2601.18091}, 
}

@inproceedings{
zhao2025skipgpt,
title={Skip{GPT}: Each Token is One of a Kind},
author={Anhao Zhao and Fanghua Ye and Yingqi Fan and Junlong Tong and Jing Xiong and Zhiwei Fei and Hui Su and Xiaoyu Shen},
booktitle={Forty-second International Conference on Machine Learning},
year={2025},
url={https://openreview.net/forum?id=d7v2iUSa9s}
}

@misc{deepseekai2026deepseekv4,
  title         = {{DeepSeek-V4}: Towards Highly Efficient Million-Token Context Intelligence},
  author        = {{DeepSeek-AI}},
  year          = {2026},
  month         = apr,
  eprint        = {2606.19348},
  archivePrefix = {arXiv},
  primaryClass  = {cs.CL},
  doi           = {10.48550/arXiv.2606.19348},
  url           = {https://arxiv.org/abs/2606.19348}
}

\clearpage 
\newpage
\appendix
\section*{Appendix}

\section{Prompting Details}
\label{appendix:prompting_details}

\subsection{Prompting Templates for Summarization}

All our summarization prompts share a common base structure. We generate prompts by populating slots within this template:

\begin{figure}[htbp]
    \begin{tcolorbox}[
        colback=blue!5!white,
        colframe=blue!75!black,
        title=The base structure of summary prompt,
        fonttitle=\small\bfseries, 
        fontupper=\footnotesize,
        top=2mm, 
        bottom=2mm,
        left=2mm,
        right=2mm,
        boxsep=1mm, 
    ]
TASK: Prune the following THOUGHT PROCESS. This is a high-difficulty task requiring careful consideration and strict adherence to all constraints.

\medskip 

STRICT LIMIT:

\texttt{\{STRICT\_LIMIT\_SECTION\}}

\medskip

COMPRESS STRATEGY:

\texttt{\{COMPRESS\_STRATEGY\_SECTION\}}

\medskip

VERIFICATION: After your summary, report the exact word count.

\medskip

QUESTION: 

\texttt{\{question}\}

\medskip

THOUGHT PROCESS: 

\texttt{\{answer}\}

\medskip

Format your response EXACTLY like this:

[The compressed thought process]

Words: [Number of words in your summary]
    \end{tcolorbox}
    \caption{The base structure of the prompt used for summarization-based methods. Different sections are populated based on the desired level of structural transformation.}
    \label{fig:prompt-base} 
\end{figure}

\subsection{Component 1: Degree of Structural Transformation}
The \texttt{\{COMPRESS\_STRATEGY\_SECTION\}} in our base prompt is populated with instructions corresponding to the four levels of structural transformation defined in the main paper.

\paragraph{Level 1: Structural Preservation (SP).}
The prompt instructs the model to only remove redundant tokens without altering the logical structure.

\begin{tcolorbox}[
colback=blue!5!white,
        colframe=blue!75!black,
        title=The SP Compress Strategy Section,
        fonttitle=\small\bfseries, 
        fontupper=\footnotesize,
        top=2mm, 
        bottom=2mm,
        left=2mm,
        right=2mm,
        boxsep=1mm, 
] 
- For this task, "pruning" means you should only remove redundant tokens.

- The goal is to preserve the original structure and sequence of every logical step.

- This strategy explicitly forbids rewriting sentences, merging steps, or deleting any part of the reasoning path.

- Conclude with the "ANSWER" section only for questions with a definitive, boxable answer. Otherwise, omit it entirely.
\end{tcolorbox}

\paragraph{Level 2: Structural Editing (SE).}
The prompt grants the model step-level editing capabilities, including keeping, deleting, and rewriting/merging steps.

\begin{tcolorbox}[
colback=blue!5!white,
        colframe=blue!75!black,
        title=The SE Compress Strategy Section,
        fonttitle=\small\bfseries, 
        fontupper=\footnotesize,
        top=2mm, 
        bottom=2mm,
        left=2mm,
        right=2mm,
        boxsep=1mm, 
]
You are an expert in logical reasoning compression. Your task is to refine and compress the provided THOUGHT PROCESS by treating it as a sequence of implicit logical steps and applying the following revision actions:

1.  **Evaluate each logical component/step** within the THOUGHT PROCESS.

2.  Based on your evaluation, decide how to act on it:

    *   **KEEP:** If a step is necessary and concise, keep it.
    
    *   **DELETE:** If a step is redundant, unnecessary, or logically flawed, remove it.
    
    *   **SINGLE-STEP COMPRESS:** If a step is necessary but verbose, rewrite it more concisely.
    
    *   **MULTI-STEP COMPRESS:** If several adjacent steps can be logically combined into a single, more efficient step, merge them.
    
3.  Synthesize your revisions into a single, coherent, and fluent reasoning path that preserves the core logical progression to the final answer. Ensure the final answer is clearly stated at the end.

4.  Conclude with the "ANSWER" section only for questions with a definitive, boxable answer. Otherwise, omit it entirely.
\end{tcolorbox}

\paragraph{Level 3: Semantic Fusion (SF).}
The prompt encourages abstraction by instructing the model to retain only the essential "key info."

\begin{tcolorbox}[
colback=blue!5!white,
        colframe=blue!75!black,
        title=The SF Compress Strategy Section,
        fonttitle=\small\bfseries, 
        fontupper=\footnotesize,
        top=2mm, 
        bottom=2mm,
        left=2mm,
        right=2mm,
        boxsep=1mm, 
]
- Retain the key info needed to solve the question and do not add additional 
info that is not included in the origin.

- Conclude with the "ANSWER" section only for questions with a definitive, boxable answer. Otherwise, omit it entirely.
\end{tcolorbox}

\paragraph{Level 4: Framework Replacement (FR).}
The prompt for FR marks a conceptual shift from "compression" to "reconstruction." This is explicitly signaled to the model by altering keywords in the base template itself. Specifically, the main \texttt{TASK} description is changed to \textbf{"Re-evaluate and Reconstruct..."} and the strategy section is retitled \texttt{RECONSTRUCTION STRATEGY}. This directs the model to abandon the original CoT's structure and generate a new reasoning trace from first principles.

\begin{tcolorbox}[
colback=blue!5!white,
        colframe=blue!75!black,
        title=The FR Reconstruction Strategy,
        fonttitle=\small\bfseries, 
        fontupper=\footnotesize,
        top=2mm, 
        bottom=2mm,
        left=2mm,
        right=2mm,
        boxsep=1mm, 
]
1.  **PRIMARY GOAL:** Your main goal is to **REPLACE** the original `THOUGHT PROCESS`. You are **NOT** compressing or editing the original text. You are generating a superior, structured replacement from scratch.\\

2.  **ASSESS COMPLEXITY:** First, analyze the provided `QUESTION` to assess its complexity (simple, moderately complex, or highly complex).\\

3.  **CHOOSE FRAMEWORK:** Based on your assessment, choose **ONE** of the following three frameworks to generate a new thought process. \\

4.  **FRAMEWORKS:**\\

    *   **For simple problems:**
        **Analysis:** [Understand the core elements and goals of the problem]
        
        **Approach:** [Propose direct solution methods]
        
        **Summary:** [Concisely summarize the solution approach and key points]\\

    *   **For moderately complex problems:**
        **Analysis:** [Understand the problem and identify key points and challenges]
        
        **Initial Approach:** [Propose preliminary solutions]
        
        **Reflection:** [Evaluate the pros and cons of the initial approach]
        
        **Improvement:** [Refine the solution based on reflection]
        
        **Summary:** [Summarize the key points of the final solution]\\

    *   **For highly complex problems:**
    
        **Analysis:** [Analyze various aspects of the problem and break down its structure]
        
        **Problem Decomposition:** [Break complex problem into manageable sub-problems]
        
        **Sub-problem Processing:** (Repeat for each sub-problem)
        
        - Sub-problem X: [Initial approach, Reflection, Improved solution]
        
        **Integration:** [Integrate sub-problem solutions]
        
        **Overall Reflection:** [Evaluate the integrated solution]
        
        **Final Optimization:** [Make final optimizations]
        
        **Summary:** [Summarize key points]\\

5.  **REFERENCE USAGE:** Use the original `THOUGHT PROCESS` (`{answer}`) only as a reference to understand the key values, steps, and final conclusion. **DO NOT** follow its structure, wording, or verbosity. Your new output must be self-contained and strictly adhere to the selected structured framework.
\end{tcolorbox}

\subsection{Component 2: Compression Ratio Control}

The \texttt{\{STRICT\_LIMIT\_SECTION\}} component controls the target length of the compressed output. We use two modes.

\paragraph{Fixed-Ratio Mode.}
Used to enforce a specific compression ratio $\gamma$. The word limit is pre-calculated based on the original CoT's length.

\begin{tcolorbox}[
colback=blue!5!white,
        colframe=blue!75!black,
        title=The FR Compress Strategy Section,
        fonttitle=\small\bfseries, 
        fontupper=\footnotesize,
        top=2mm, 
        bottom=2mm,
        left=2mm,
        right=2mm,
        boxsep=1mm, 
]
STRICT LIMIT: Your summary MUST be around \texttt{\{word limit}\}  words. This is a non-negotiable hard limit.
\end{tcolorbox}

\paragraph{Self-Determined Mode.}
Used to let the model determine the optimal length based on the compression strategy itself, which is useful for establishing a natural compression baseline for each strategy.

\begin{tcolorbox}[
colback=blue!5!white,
        colframe=blue!75!black,
        title=The FR Compress Strategy Section,
        fonttitle=\small\bfseries, 
        fontupper=\footnotesize,
        top=2mm, 
        bottom=2mm,
        left=2mm,
        right=2mm,
        boxsep=1mm, 
]
STRICT LIMIT:\\
   \#\#\# Phase 1: Internal Estimation (Do this in your mind)\\
   1. First, carefully read the entire THOUGHT PROCESS.\\
   2. Then, mentally apply the rules from the COMPRESS STRATEGY below.\\
   3. Estimate the final word count that would result from this pruning process. Let's call this your "Self-Determined Budget".\\

   \#\#\# Phase 2: Execution\\
   1. Now, perform the actual pruning according to the COMPRESS STRATEGY.\\
   2. Your final output's word count MUST be close to the "Self-Determined Budget" you estimated in Phase 1.
\end{tcolorbox}

\section{Condition-Aware Decision Guide}
\label{app:decision_guide}

Table~\ref{tab:decision_guide} translates our empirical findings into
practical choices within selective pruning and generative rewriting.
It does not assume that either method family is universally superior.

\begin{table*}[t]
\centering
\footnotesize
\setlength{\tabcolsep}{4pt}
\renewcommand{\arraystretch}{1.12}

\begin{tabularx}{\textwidth}{
    @{}
    >{\raggedright\arraybackslash}p{3.5cm}
    >{\raggedright\arraybackslash}X
    >{\raggedright\arraybackslash}X
    @{}
}
\toprule
\textbf{Condition}
& \textbf{Recommended choice}
& \textbf{Evidence and caveat} \\
\midrule

\rowcolor{groupgray}
\multicolumn{3}{@{}l}{\textbf{Selective pruning}} \\

Reasoning traces can be reliably segmented into steps
& Use step-level pruning; the choice of importance criterion is secondary.
& Different criteria converge on a shared reasoning backbone and yield
similar downstream accuracy (Section~4.3). \\

Fine-grained token pruning is required
& Use LLMLingua-2 or explicitly protect operators, numbers, constraints,
and intermediate values.
& Unprotected Entropy/NLL pruning may remove symbolic anchors, especially
under aggressive compression on Math tasks (Sections~4.2--4.3). \\

\addlinespace[2pt]
\rowcolor{groupgray}
\multicolumn{3}{@{}l}{\textbf{Generative rewriting}} \\

Symbolic or math-heavy tasks
& Use structural preservation (SP) with a conservative compression budget.
& Math accuracy degrades monotonically as restructuring increases from
SP to FR (Section~5.2). \\

Knowledge-intensive, general-domain tasks
& Consider semantic fusion (SF), or FR with explicit validation, when
stronger compression is desired; use SP when stability is the priority.
& Aggressive rewriting can match or exceed SP on General tasks, but FR
exhibits higher variance (Sections~5.1--5.2). \\

\addlinespace[2pt]
\rowcolor{groupgray}
\multicolumn{3}{@{}l}{\textbf{Compression budget}} \\

Unclear rewriting budget 
& Choose the restructuring level first, start from the free-ratio setting,
then measure and validate the realized compression ratio.
& Structural constraints determine the achievable length band; the target
ratio is not a precise length controller (Section~6.2). \\

Aggressive compression is required
& Sweep and validate nearby ratios rather than extrapolating from mild
compression.
& Accuracy follows a plateau-then-drop pattern; Math tasks generally reach
the failure region earlier (Section~6.1). \\

\addlinespace[2pt]
\rowcolor{groupgray}
\multicolumn{3}{@{}l}{\textbf{Deployment}} \\

The student is trained with Short-CoT supervision
& Use the realized training ratio as an approximate inference-cost proxy,
but verify it on the target model.
& Training and inference ratios are closely aligned in the Short-CoT
regime (Section~6.3). \\

The student is trained with Long-CoT supervision
& Measure inference length directly against the deployment budget.
& Students may retain verbose reasoning habits; the training ratio is an
optimistic lower bound on inference cost (Section~6.3). \\

\bottomrule
\end{tabularx}

\caption{A condition-aware decision guide derived from Sections~4--6.
The recommendations describe empirically supported defaults within the
models, domains, and compression regimes evaluated in this work.}
\label{tab:decision_guide}
\end{table*}
\section{Experimental Details}
\label{appendix:exp_details}


\begin{table*}[ht!]
\centering
\small
\caption{Core fine-tuning hyperparameters for short and long-CoT regimes. The long-CoT regime uses a lower learning rate, more epochs due to a smaller dataset, and a larger effective batch size to ensure stable and robust learning.}
\label{tab:training_configs}
\begin{tabular}{lcc}
\toprule
\textbf{Parameter} & \textbf{Short-CoT Regime} & \textbf{Long-CoT Regime} \\
\midrule
\multicolumn{3}{l}{\textit{LoRA Configuration}} \\
Rank ($r$) & 16 & 16 \\
Alpha ($\alpha$) & 32 & 32 \\
Dropout & 0.05 & 0.05 \\
\midrule
\multicolumn{3}{l}{\textit{Training Dynamics}} \\
Learning Rate & 5e-5 & 1e-5 \\
LR Scheduler & Cosine & Cosine \\
Warmup Ratio & 0.1 & 0.05 \\
Num. Train Epochs & 3 & 4 \\
Optimizer & AdamW & AdamW (via DeepSpeed ZeRO-2) \\
Effective Batch Size & 64 & 64 \\
\quad \textit{(per-dev bs × grad accum × \#gpus)} & \textit{(4 × 16 × 1)} & \textit{(1 × 32 × 2)} \\
\bottomrule
\end{tabular}
\end{table*}

\subsection{Implementation Details}
\label{appendix:training_details}
\paragraph{Training Corpora. ($\mathcal{R}$)} To study how compression strategies perform across different reasoning lengths, we create training sets from two representative datasets. For the \textit{short-CoT} regime, we use 5,000 instances from the Natural Reasoning dataset~\citep{yuan2025naturalreasoningreasoningwild28m}, which features compact reasoning traces. For the \textit{long-CoT} regime, we use the LIMO dataset~\citep{ye2025limoreasoning}, which contains extended, multi-step rationales. This setup allows us to analyze performance in both data-abundant (short) and data-scarce (long) but highly detailed reasoning scenarios.

\paragraph{Fine-tuning Configuration.}
We fine-tune all student models using Parameter-Efficient Fine-Tuning (PEFT), specifically LoRA~\citep{lora}, managed by the \texttt{LLaMA-Factory}~\citep{llamafactory} framework. Key hyperparameters, which were optimized for either short or long context regimes, are summarized in Table~\ref{tab:training_configs}. For both LLaMA-3.1 and Qwen2.5 models, LoRA was applied to a comprehensive set of layers, including all attention-related projections (\texttt{q\_proj, k\_proj, v\_proj, o\_proj}) and feed-forward network layers (\texttt{gate\_proj, up\_proj, down\_proj}), to maximize adaptive capacity.

Training was performed using \texttt{bfloat16} mixed-precision. To handle the extensive sequence lengths of the long-CoT regime and improve training efficiency, we employed FlashAttention-2~\citep{dao2024flashattention2} and the DeepSpeed ZeRO-2 optimizer~\citep{rajbhandari2020zero}. The maximum sequence length (\texttt{cutoff\_len}) was set to 2048 for short-CoT data and 16,384 for long-CoT data.

\subsection{Evaluation}
\paragraph{Evaluation Benchmarks ($d$)} To assess the acquired reasoning capabilities across diverse domains, we evaluate on five challenging benchmarks:
\begin{itemize}[leftmargin=*, topsep=0pt, itemsep=2pt]
\item \textbf{Mathematical Reasoning}: GSM8K~\citep{Cobbe:2021gsm8k}, MATH500~\citep{math}, AMC23~\citep{maa_amc_2025}and the high-school competition-level AIME24~\citep{aops2025aimeproblems}.
\item \textbf{General Reasoning}: GPQA Diamond~\citep{rein2024gpqa} (graduate-level science) and MMLU-PRO~\citep{wang2024mmlupro} (academic exams).
\end{itemize}

\paragraph{Inference Settings}
We run all evaluations with the vLLM backend. The backend is configured as \texttt{tensor\_parallel\_size=1}, \texttt{dtype=bfloat16}, \texttt{max\_model\_len=32768}, and \texttt{gpu\_memory\_utilization=0.9}, with optional LoRA adapters passed through \texttt{lora\_adapter}. Sampling uses \texttt{temperature=0.7}, \texttt{top\_p=1.0}, \texttt{seed=42}, and \texttt{n=1}. The maximum generation length is task-specific: \texttt{8192} tokens for GSM8K, MATH500, GPQA-Diamond, MMLU-Pro, and AMC23, and \texttt{15000} tokens for AIME24. All runs are executed on a single NVIDIA A100 40GB GPU with CUDA 12.9.1 and 8 CPU cores (64GB RAM).

\paragraph{Evaluation Metrics.} We jointly measure reasoning quality and inference efficiency. All evaluations use a standardized decoding configuration (temperature=0.7, top-p=1.0) to ensure fair comparison across all models and strategies.
\begin{itemize}[leftmargin=*, topsep=0pt, itemsep=2pt]
\item \textbf{Accuracy:} We report the percentage of problems where the model's final answer matches the ground truth for performance ($\mathrm{Perf}$) 
\item \textbf{Generated Length:} We report the average number of tokens in the model's generated response (CoT + answer) as the inference Token Ratio ($\kappa$) for cost.
\end{itemize}
Full implementation, hyperparameter, and evaluation details are provided in the Appendix~\ref{appendix:training_details}.
\section{Method Details}
\label{appendix: method}

\subsection{Extraction-based Pruning}

Extraction methods prune a CoT, $C$, by following a three-stage pipeline: 1) Scoring: computing an importance score $I(u)$ for each unit $u \in C$; 2) Thresholding: deriving a pruning threshold $I_\gamma$ from the scores based on a target ratio $\gamma$; and 3) Pruning: removing units with $I(u) < I_\gamma$. The core of this paradigm is the choice of the scoring function, which we term the redundancy signal.

\paragraph{Importance criterion.} 
Each unit $u$ (a word or a step) receives a score $s(u)$, where larger values indicate higher importance (more worth preserving).
\begin{itemize}[leftmargin=*, topsep=0pt, itemsep=2pt]
\item \textbf{LLMLingua-2 (preserve probability):}
$I(u)$ is the probability that a unit should be preserved predicted by a bidirectional scorer $M_B$:
\begin{equation}
I_{\text{lingua}}(t_i)=P\!\left(y_i=\mathrm{preserve}\mid \mathbf{t};\theta_{M_B}\right)
\end{equation}

\item \textbf{Uncertainty (Entropy):}
$I(u)$ downweights uncertain tokens, using the negative average token entropy under a language model:
\begin{gather}
H_j = -\sum_{v\in\mathcal{V}} P(v \mid u_{<j}, \mathcal{C}) \log P(v \mid u_{<j} \mathcal{C})\\
I_{\text{entropy}}(u) = \frac{1}{|u|}\sum_{j=1}^{|u|} H_j
\end{gather}
where $\mathcal{C}$ denotes the conditioning context.

\item \textbf{Predictability (NLL / Surprisal):}
$I(u)$ favors predictable units via negative average surprisal:
\begin{equation}
I_{\text{NLL}}(u)=\frac{1}{|u|}\sum_{j=1}^{|u|}\Big(-\log P(u_j\mid u_{<j},\mathcal{C})\Big)
\end{equation}
\end{itemize}

\paragraph{Unit aggregation (word/step).}
For a unit $u$ consisting of tokens $\{t_j\}_{j=1}^{|u|}$, we use the mean token importance:
\begin{equation}
I(u) \;=\; \frac{1}{|u|}\sum_{j=1}^{|u|} I(t_j),
\end{equation}
which is applied to word units and step units respectively.


\paragraph{Pruning Logic.}
Let $\gamma\in(0,1]$ denote the target compression ratio (kept fraction), and define the drop ratio $\rho = 1-\gamma$.
We assign each unit an importance score $I(\cdot)$, where larger values indicate higher importance.
We compute a threshold $I_\gamma$ using an empirical percentile with a small offset:
\begin{equation}
I_\gamma = \mathrm{Percentile}\big(\{I(u_i)\}_{i=1}^N,\ 100\rho + 1\big).
\end{equation}
Units are retained if their importance is strictly larger than the threshold:
\begin{equation}
\tilde{C} = (u_i \mid I(u_i) > I_\gamma)_{i=1}^N .
\end{equation}
We form $\tilde{C}$ by keeping all retained units in their original order (i.e., an order-preserving subsequence of $C$).
For token-level compression we first aggregate token scores into word scores (mean over tokens within each word) and apply the same percentile rule over words;  For step-level compression we apply the percentile rule directly over step scores.

\subsection{Summarization-based Rewriting}

Summarization methods leverage a powerful LLM Gemini2.5-Flash~\citep{comanici2025gemini25pushingfrontier} as a "rewriter" to generate a shortened trajectory $\tilde{r}$, as shown in Eq. ~\ref{eq:rewrite}. To navigate this design space, we introduce the Degree of Restructing Level, a novel dimension that categorizes rewriting strategies based on how deeply they alter the original CoT's logic. It spans four ordered levels:

\begin{itemize}[leftmargin=*, itemsep=2pt]
\item \textbf{L1: Structural Preservation (SP):} The rewriter is constrained to only remove redundant tokens, preserving the original sequence of logical steps verbatim and targeting only surface-level verbosity.
\item \textbf{L2: Structural Editing (SE):} The rewriter performs step-level operations like merging or rephrasing to create a more efficient but structurally similar reasoning path.
\item \textbf{L3: Semantic Fusion (SF):} The rewriter is encouraged to retain only "key information," allowing it to abstract a multi-step sequence into a single, novel statement, achieving a higher level of abstraction.
\item \textbf{L4: Framework Replacement (FR):} The rewriter discards the original structure entirely and regenerates the reasoning process following a predefined formal template, enforcing a complete structural overhaul.
\end{itemize}
The specific prompts used to elicit these behaviors are detailed in Appendix ~\ref{appendix:prompting_details}.

\subsection{Compression Budget}
\paragraph{Compression ratios.}
For each example $i$, let $L_i^{\text{teacher}}$ denote the token length of the original (uncompressed) teacher CoT, and $L_i^{\text{comp}}$ the token length of the compressed CoT produced by a given method. We control and measure compression through three related ratios, computed at the instance level and then averaged over the dataset.

\textbf{Target ratio.}
We denote by $\rho_{\text{target}} \in (0, 1]$ the desired compression level specified during training or prompting. Intuitively, $\rho_{\text{target}}$ corresponds to the intended fraction of teacher tokens to keep:
\begin{equation}
    \frac{\sum_{i=1}^N L_i^{\text{comp}}}{\sum_{i=1}^N L_i^{\text{teacher}}} \approx \rho_{\text{target}}.
\end{equation}
For each strategy, we train variants at three target ratios $\gamma \in \{0.9, 0.7, 0.5\}$.

\textbf{Realized ratio.}
The realized ratio $\rho_{\text{inference}}$ is the actual compression achieved by the method:
\begin{equation}
    \rho_{\text{realized}} =\; \frac{1}{N} \sum_{i=1}^N \frac{L_i^{\text{comp}}}{L_i^{\text{teacher}}}.
\end{equation}
This is the quantity we use on the horizontal axis when relating compression strength to accuracy.

\textbf{Inference token ratio.}
At evaluation time, we also track the end-to-end token cost of answering benchmark questions. For each test example $i$, let $T_i^{\text{meth}}$ be the number of tokens in the CoT-plus-final-answer produced by a compressed method, and $T_i^{\text{teacher}}$ the same quantity under the original (uncompressed) teacher-CoT setting. The inference token ratio $\kappa$ is defined as
\begin{equation}
    \rho_{\text{inference}} =\; 
    \frac{\frac{1}{N} \sum_{i=1}^N T_i^{\text{meth}}}
         {\frac{1}{N} \sum_{i=1}^N T_i^{\text{teacher}}}.
\end{equation}
This is the quantity of direct interest for deployment-time token budgets.

\subsection{Cross-Rewriter Analysis}
\label{appendix:cross_rewriter}

Beyond Gemini 2.5-Flash, we explored GPT-4o-mini as an alternative rewriter 
to characterize how rewriter choice influences compression behavior under the 
Short-CoT regime. As shown in Table~\ref{tab:cross_rewriter}, the two rewriters 
differ in absolute compression calibration, yet the structural ordering is 
consistent: Strong fidelity yields the most conservative compression, Light 
the most aggressive, and band width grows as structural constraints loosen. 
This suggests that length band structure is primarily governed by the degree 
of structural transformation rather than the rewriter itself.

\begin{table}[h]
\centering
\small
\begin{tabular}{llcc}
\toprule
\textbf{Structure} & \textbf{Target $\gamma$} & \textbf{Gemini $\rho$} & \textbf{GPT-4o-mini $\rho$} \\
\midrule
Strong & 0.9 & 0.932 & 0.812 \\
Strong & 0.7 & 0.847 & 0.714 \\
Strong & 0.5 & 0.760 & 0.602 \\
\midrule
Medium & 0.9 & 0.940 & 0.882 \\
Medium & 0.7 & 0.782 & 0.725 \\
Medium & 0.5 & 0.574 & 0.530 \\
\midrule
Light  & 0.9 & 0.931 & 0.831 \\
Light  & 0.7 & 0.755 & 0.682 \\
Light  & 0.5 & 0.605 & 0.511 \\
\bottomrule
\end{tabular}
\caption{Realized compression ratios ($\rho$) for Gemini 2.5-Flash and GPT-4o-mini 
across structural fidelity levels and target ratios ($\gamma$) under the Short-CoT regime.}
\label{tab:cross_rewriter}
\end{table}
\section{Definitions, Pareto Construction, and Full Results}
\label{appendix:Result_equation}

\begin{table*}[t]
\centering
\small
\setlength{\tabcolsep}{1.2mm}
\begin{tabular}{@{}l c r r c r r c r r c r r c@{}}
\toprule
\multirow{3}{*}{\textbf{Methods}} & \multirow{3}{*}{\textbf{Ratio}}
& \multicolumn{6}{c}{\textbf{MATH}} & \multicolumn{6}{c}{\textbf{GENERAL}} \\
\cmidrule(lr){3-8} \cmidrule(lr){9-14}
& & \multicolumn{3}{c}{\textbf{Step}} & \multicolumn{3}{c}{\textbf{Token}}
  & \multicolumn{3}{c}{\textbf{Step}} & \multicolumn{3}{c}{\textbf{Token}} \\
\cmidrule(lr){3-5} \cmidrule(lr){6-8} \cmidrule(lr){9-11} \cmidrule(lr){12-14}
& & Acc & Tok & InferRatio$\kappa$
  & Acc & Tok & InferRatio$\kappa$
  & Acc & Tok & InferRatio$\kappa$
  & Acc & Tok & InferRatio$\kappa$ \\
\midrule

\multicolumn{14}{c}{\textbf{\textsc{NATURAL REASONING}}} \\
\addlinespace[2pt]

\texttt{Original}  & - 
& 67.1 & 528 & 67.1 
& 67.1 & 528 & 67.1 
& 35.4 & 572 & 32.6
& 35.4 & 572 & 32.6 \\

\midrule
\multirow{3}{*}{\texttt{L2}}  
& 0.9 
& 64.6 & 447 & 84.6 
& 65.2 & 466 & 88.2
& 36.9 & 537 & 93.9
& 36.3 & 441 & 77.1 \\
& 0.7 
& 59.4 & 404 & 76.6
& 61.7 & 386 & 73.1
& 39.1 & 450 & 78.7
& 35.5 & 405 & 70.9 \\
& 0.5 
& 49 & 377 & 71.5 
& 50.1 & 279 & 52.9
& 37.7 & 368 & 64.3
& 32.7 & 300 & 52.5 \\

\midrule

\multirow{3}{*}{\texttt{ENT}}  
& 0.9 
& 64.6 & 518 & 98.1 
& 63.7 & 460 & 87.2
& 35.9 & 609 & 106.5
& 34.9 & 502 & 87.8 \\
& 0.7 
& 46.3 & 395 & 74.8
& 58.3 & 356 & 67.4
& 38.9 & 499 & 87.2
& 33.8 & 391 & 68.4 \\
& 0.5 
& 31.2 & 299 & 566.6
& 46.5 & 257 & 48.7
& 34.3 & 389 & 68.1
& 30.6 & 372 & 65.1 \\

\midrule

\multirow{3}{*}{\texttt{NLL}}  
& 0.9 
& 63.2 & 523 & 99.1
& 64.0 & 489 & 92.6
& 36.0 & 628 & 109.8
& 34.2 & 582 & 101.7 \\
& 0.7 
& 42.2 & 402 & 76.2
& 58.5 & 398 & 75.4
& 36.7 & 499 & 87.2
& 35.5 & 452 & 79.1 \\
& 0.5 
& 33.3 & 313 & 59.2 
& 45 & 261 & 49.5
& 34.8 & 398 & 69.6
& 30.3 & 402 & 70.3 \\

\midrule
\multicolumn{14}{c}{\textbf{\textsc{LIMO}}} \\
\addlinespace[2pt]

\texttt{Original}  & - 
& 27.1 & 3033 & - 
& 27.1 & 3033 & - 
& 40.4 & 1454 & -
& 40.4 & 1454 & - \\

\midrule
\multirow{3}{*}{\texttt{L2}}  
& 0.9 
& 25.6 & 2596 & 85.6
& 24.3 & 2115 & 69.7
& 41.2 & 1161 & 79.8
& 39.5 & 1069 & 73.5 \\
& 0.7 
& 23.7 & 2582 & 85.1
& 23.8 & 2575 & 84.9
& 40.6 & 1278 & 87.9
& 38.2 & 1244 & 85.5 \\
& 0.5 
& 28.7 & 2461 & 81.1
& 27.2 & 2173 & 71.6
& 40.9 & 1289 & 88.6
& 39.9 & 1246 & 85.7 \\

\midrule

\multirow{3}{*}{\texttt{ENT}}  
& 0.9 
& 26.6 & 2477 & 81.7
& 26.3 & 2254 & 74.3
& 38 & 1100 & 75.6
& 38.5 & 1147 & 78.9 \\
& 0.7 
& 25.4 & 2867 & 94.5
& 24.6 & 1935 & 63.8
& 37.8 & 1126 & 77.4
& 40.5 & 1078 & 74.1 \\
& 0.5 
& 23.4 & 2116 & 69.7
& 25.9 & 1836 & 60.5
& 39.2 & 1162 & 79.9
& 38.8 & 1196 & 82.3 \\

\midrule

\multirow{3}{*}{\texttt{NLL}}  
& 0.9 
& 24.2 & 2318 & 76.4
& 30.2 & 2065 & 68.1
& 39.8 & 1102 & 75.8
& 39.7 & 1320 & 90.7 \\
& 0.7 
& 23.6 & 2390 & 78.8
& 22.0 & 1867 & 61.5
& 38.6 & 1131 & 77.8
& 40.5 & 1127 & 77.5 \\
& 0.5 
& 25.0 & 2193 & 72.3
& 27.1 & 1811 & 59.7
& 41.8 & 1078 & 74.1
& 42.0 & 1201 & 82.6 \\

\bottomrule
\end{tabular}
\caption{Experimental results of \method on LLaMA-3.1-8B-Instruct. We report accuracy(acc), average CoT token count (Tok), and actual compression ratio (InferRatio$\kappa$) for comparison. Accuracy (Acc) and the actual compression ratio (InferRatio$\kappa$) are both reported as percentages.
(L2: LLMLingua2, ENT: Entropy)}
\label{tab:llama_extraction}
\end{table*}

\begin{table*}[t]
\centering
\small
\setlength{\tabcolsep}{1.2mm}
\begin{tabular}{@{}l c r r c r r c r r c r r c@{}}
\toprule
\multirow{3}{*}{\textbf{Methods}} & \multirow{3}{*}{\textbf{Ratio}}
& \multicolumn{6}{c}{\textbf{MATH}} & \multicolumn{6}{c}{\textbf{GENERAL}} \\
\cmidrule(lr){3-8} \cmidrule(lr){9-14}
& & \multicolumn{3}{c}{\textbf{Step}} & \multicolumn{3}{c}{\textbf{Token}}
  & \multicolumn{3}{c}{\textbf{Step}} & \multicolumn{3}{c}{\textbf{Token}} \\
\cmidrule(lr){3-5} \cmidrule(lr){6-8} \cmidrule(lr){9-11} \cmidrule(lr){12-14}
& & Acc & Tok & InferRatio$\kappa$
  & Acc & Tok & InferRatio$\kappa$
  & Acc & Tok & InferRatio$\kappa$
  & Acc & Tok & InferRatio$\kappa$ \\
\midrule

\multicolumn{14}{c}{\textbf{\textsc{NATURAL REASONING}}} \\
\addlinespace[2pt]
\texttt{Original}  & - 
& 53.1 & 507 & -
& 53.1 & 507 & -
& 33.6 & 537 & -
& 33.6 & 537 & - \\

\midrule
\multirow{3}{*}{\texttt{L2}}  
& 0.9 
& 50.3 & 442 & 87.3
& 51.2 & 425 & 83.8
& 32.7 & 505 & 94
& 30.9 & 447 & 83.1 \\
& 0.7 
& 40.5 & 502 & 99.1
& 48.2 & 355 & 70.1
& 33.2 & 618 & 115
& 31.1 & 378 & 70.3 \\
& 0.5 
& 25.3 & 821 & 162.1
& 43.5 & 279 & 55.1
& 35 & 593 & 110.3
& 31.4 & 328 & 61 \\

\midrule

\multirow{3}{*}{\texttt{ENT}}  
& 0.9 
& 47.2 & 447 & 88.3 
& 48 & 421 & 83
& 29 & 500 & 93.1
& 31.6 & 409 & 76.2 \\
& 0.7 
& 27.3 & 381 & 75.2
& 42.5 & 364 & 71.8
& 29.7 & 480 & 89.4
& 31.7 & 358 & 66.6 \\
& 0.5 
& 17.7 & 332 & 65.4
& 36.1 & 300 & 59.1
& 29.7 & 405 & 75.3
& 35.4 & 266 & 49.4 \\

\midrule

\multirow{3}{*}{\texttt{NLL}}  
& 0.9 
& 42.5 & 446 & 88.1
& 47.3 & 433 & 85.5
& 31.7 & 548 & 102
& 30.8 & 397 & 73.8 \\
& 0.7 
& 25.8 & 365 & 72
& 42.4 & 381 & 75.2
& 31.4 & 527 & 98
& 33.9 & 348 & 64.7 \\
& 0.5 
& 17.3 & 222 & 43.8
& 37 & 304 & 59.9
& 30.8 & 403 & 75
& 36.1 & 280 & 52.1 \\

\midrule
\multicolumn{14}{c}{\textbf{\textsc{LIMO}}} \\
\addlinespace[2pt]
\texttt{Original}  & - 
& 27 & 1368 & -
& 27 & 1368 & -
& 38.9 & 537 & -
& 38.9 & 537 & - \\

\midrule
\multirow{3}{*}{\texttt{L2}}  
& 0.9 
& 26.9 & 1053 & 77
& 25.6 & 869 & 63.5
& 40.4 & 499 & 92.8
& 40.1 & 547 & 101.8 \\
& 0.7 
& 25.9 & 1026 & 75
& 24 & 998 & 72.9
& 39.1 & 535 & 99.6
& 39.4 & 553 & 102.9 \\
& 0.5 
& 23.3 & 1137 & 83.1
& 25.5 & 1151 & 84.1
& 35.7 & 545 & 101.4
& 37 & 548 & 102 \\

\midrule

\multirow{3}{*}{\texttt{ENT}}  
& 0.9 
& 23.7 & 1262 & 92.2
& 24.2 & 899 & 65.7
& 40.7 & 541 & 100.7
& 38.4 & 572 & 106.5 \\
& 0.7 
& 25.8 & 1079 & 78.9
& 26.2 & 1023 & 74.8
& 40.3 & 564 & 104.9
& 40.5 & 585 & 108.8 \\
& 0.5 
& 26 & 981 & 71.7
& 24 & 813 & 59.4
& 41.7 & 562 & 104.7
& 40.9 & 571 & 106.3 \\

\midrule

\multirow{3}{*}{\texttt{NLL}}  
& 0.9 
& 25.7 & 1137 & 83.1
& 26.2 & 953 & 69.6
& 38.7 & 553 & 102.9
& 39.7 & 513 & 95.5 \\
& 0.7 
& 24 & 886 & 64.8
& 24 & 850 & 62.1
& 38.8 & 556 & 103.4
& 40.1 & 542 & 100.8 \\
& 0.5 
& 26.1 & 900 & 65.8
& 24.3 & 1008 & 73.7
& 40.1 & 543 & 101
& 39.1 & 544 & 101.2 \\

\bottomrule
\end{tabular}
\caption{Experimental results of \method on Qwen2.5-1.5B-Instruct. We report accuracy(acc), average CoT token count (Tok), and actual compression ratio (InferRatio$\kappa$) for comparison. Accuracy (Acc) and the actual compression ratio (InferRatio$\kappa$) are both reported as percentages.
(L2: LLMLingua2, ENT: Entropy)}
\label{tab:qwen_extraction}
\end{table*}

\begin{table*}[t]
\centering
\small
\setlength{\tabcolsep}{1.1mm} 
\begin{tabular}{@{}lc*{4}{rrc}@{}}
\toprule
\multirow{3}{*}{\textbf{Methods}} & \multirow{3}{*}{$\textbf{Ratio}$}
& \multicolumn{6}{c}{\textbf{LLaMA-3.1-8B-Instruct}}
& \multicolumn{6}{c}{\textbf{Qwen2.5-7B-Instruct}} \\
\cmidrule(lr){3-8} \cmidrule(lr){9-14}
& & \multicolumn{3}{c}{\textbf{MATH}}  & \multicolumn{3}{c}{\textbf{GENERAL}}
  & \multicolumn{3}{c}{\textbf{MATH}}  & \multicolumn{3}{c}{\textbf{GENERAL}} \\
\cmidrule(lr){3-5} \cmidrule(lr){6-8} \cmidrule(lr){9-11} \cmidrule(lr){12-14}
& & Acc & Tok & InferRatio$\kappa$
  & Acc & Tok & InferRatio$\kappa$
  & Acc & Tok & InferRatio$\kappa$
  & Acc & Tok & InferRatio$\kappa$ \\
\midrule

\texttt{Original}  & - 
& 67.1 & 528 & 61.7 & 35.4 & 572 & 32.6
& 53.1  & 507  & --   & 33.6  & 537  & -- \\ \midrule

\multicolumn{14}{c}{\textbf{\textsc{NATURAL REASONING}}} \\
\addlinespace[2pt]
\multirow{4}{*}{\texttt{SP}}
&0.9 
& 66.9 & 522 & 98.9 & 39.8 & 582.5 & 101.9
& 50.6 & 464 & 91.6 & 397  & 533.5 & 99.3 \\
&0.7 
& 66.2 & 564 & 106.8 & 37.7 & 547 & 95.7
& 48.9 & 432 & 85.2 & 40.4 & 208 & 38.7 \\
&0.5 
& 68.1 & 442 & 83.7 & 38.2 & 465 & 81.4
& 49.8 & 411 & 81 & 41  & 389.5  & 72.5 \\
&0.35
& 66.8 & 482 & 91.4 & 34.7 & 541 & 94.7
& 50.8 & 457 & 90.2 & 30.3 & 496.5 & 92.5 \\ \midrule

\multirow{4}{*}{\texttt{SE}}
&0.9 
& 59 & 461 & 87.3 & 42.2 & 537.5 & 94.1
& 48.1 & 319 & 63 & 45.4 & 406.5 & 75.7 \\
&0.7 
& 61.5 & 358 & 67.9 & 41.6 & 421.5 & 73.8
& 47.9 & 262 & 51.6 & 49.2 & 339.5 & 63.2 \\
&0.5 
& 58.8 & 327 & 62 & 40.6 & 341 & 59.7
& 48.8 & 230 & 45.3 & 487 & 202.5 & 37.7 \\
&0.35 
& 58.1 & 265 & 50.1 & 32.4 & 241 & 42.2
& 46.9 & 174 & 34.3 & 45 & 134.5 & 25 \\ \midrule

\multirow{4}{*}{\texttt{SF}}
&0.9 
& 61.8 & 422 & 0.8 & 43 & 532 & 93.1
& 49.9 & 370 & 73.1 & 47.4 & 530 & 98.7 \\
&0.7 
& 60.7 & 438 & 82.9 & 24.8 & 412.5 & 72.2
& 48.9 & 294 & 58 & 42.7 & 211.5 & 39.4 \\
&0.5 
& 58 & 413 & 78.3 & 22.9 & 391.5 & 68.5
& 49.3 & 237 & 46.8 & 43.3 & 143.5 & 26.7 \\
&0.35 
& 59.3 & 193 & 36.6 & 34.2 & 179 & 31.3
& 44.4 & 157 & 30.9 & 49 & 86 & 16 \\ \midrule

\texttt{FR}
&0.5 
& 49.4 & 341 & 64.6 & 43.4 & 421 & 73.7
& 27.5 & 270 & 53.2 & 46.4 & 336.5 & 62.7 \\ \midrule

\multicolumn{14}{c}{\textbf{\textsc{LIMO}}} \\
\addlinespace[2pt]
\multirow{4}{*}{\texttt{SP}}
&0.9 
& 24 & 2333 & 76.9 & 41.8 & 1343 & 92.4
& 23.7 & 1066 & 77.9 & 38.6 & 577 & 107.4 \\
&0.7 
& 21 & 2629 & 86.7 & 37.2 & 1248 & 85.8
& 27.4 & 1061 & 77.5 & 38.7 & 552 & 102.8 \\
&0.5 
& 20.9 & 3439 & 113.4 & 39.9 & 1290 & 88.7
& 21.9 & 1118 & 81.7 & 38.8 & 599 & 111.5 \\
&0.35
& 27 & 3069 & 101.2 & 40.8 & 1308 & 89.9
& 23.1 & 1333 & 97.4 & 39.8 & 583 & 108.6 \\ \midrule

\multirow{4}{*}{\texttt{SE}}
&0.9 
& 24.6 & 3153 & 103.9 & 37.9 & 1487 & 102.3
& 21.3 & 1494 & 109.2 & 39.1 & 595 & 110.8 \\
&0.7 
& 22.3 & 3287 & 108.1 & 41.9 & 1317 & 90.5
& 24.9 & 1210 & 88.5 & 39.4 & 608 & 113.2 \\
&0.5 
& 23.9 & 3690 & 121.6 & 38.2 & 1543 & 106.1
& 23 & 1303 & 95.2 & 39.6 & 582 & 108.4 \\
&0.35 
& 22.6 & 3068 & 101.1 & 35.9 & 896 & 61.6
& 19.3 & 1118 & 81.7 & 43.3 & 480 & 89.4 \\ \midrule

\multirow{4}{*}{\texttt{SF}}
&0.9 
& 24.1 & 3126 & 104.2 & 39.7 & 1528 & 105.1
& 25.3 & 1197 & 87.5 & 41.7 & 610 & 113.6 \\
&0.7 
& 21.8 & 2474 & 81.6 & 43.2 & 1456 & 100.1
& 24 & 1199 & 87.6 & 39.3 & 640 & 119.1 \\
&0.5 
& 21.2 & 3987 & 131.4 & 37.4 & 1618 & 111.3
& 24 & 1322 & 96.6 & 41.3 & 569 & 105.9 \\
&0.35 
& 26.3 & 1844 & 60.8 & 36.2 & 741 & 51
& 18.1 & 96.8 & 70.8 & 40.7 & 434 & 80.8 \\ \midrule

\texttt{FR}
&0.5 
& 23.2 & 2375 & 78.3 & 38.8 & 1364 & 93.8
& 23.3 & 1051 & 76.8 & 43.9 & 566 & 105.4 \\
\bottomrule
\end{tabular}
\caption{Experimental results of \method on LLaMA-3.1-8B-Instruct and Qwen2.5-1.5B-Instruct. We report accuracy (Acc), average CoT token count (Tok), and actual compression ratio (InferRatio$\kappa$) for comparison. Accuracy (Acc) and the actual compression ratio (InferRatio$\kappa$) are both reported as percentages.
}
\label{tab:llama_qwen_summary}
\end{table*}

This appendix details the formal metrics used to characterize the safe regime, the construction of the Pareto efficiency frontiers, and the complete numerical tabulations for all experiments.

\subsection{Effectiveness Criterion and Safe Regime}
\label{appendix:Effectiveness_rule}
Let $A_{\text{teacher}}$ denote the accuracy of the uncompressed teacher model, and let $A(r)$ denote the student accuracy achieved at a realized compression ratio $r$.
We define a compression level as \textit{effective} if it retains at least 90\% of the teacher's performance:
\begin{equation}
A(r) \ge 0.9 \cdot A_{\text{teacher}}.
\end{equation}
The \textbf{safe regime} is defined as the maximal contiguous interval of realized ratios starting from $r=1.0$ (no compression) for which this condition holds.
All token counts reported ($T$) include both the input prompt and the generated reasoning tokens, using the same tokenizer and decoding strategy across all comparisons.

\subsection{Pareto Frontier Construction}
To characterize the trade-off between deployment cost and performance, we analyze the Pareto frontier in the $(\text{Cost}, \text{Performance})$ plane.
Let $\mathcal{C}$ be the set of all experimental configurations (points). A point $i \in \mathcal{C}$ with inference token count $T_i$ and accuracy $A_i$ is considered \textbf{Pareto-optimal} if there exists no other point $j \in \mathcal{C}$ such that:
\[
T_j \le T_i \quad \text{and} \quad A_j \ge A_i,
\]
with at least one inequality being strict.
Practically, this frontier is constructed by sorting all valid candidates by ascending token count and retaining only those points that establish a new strictly increasing maximum in accuracy. This frontier represents the upper bound of efficiency—the maximum accuracy achievable for a given inference budget.

\subsection{Full Numerical Results}
We provide the complete numerical data underpinning our analysis in Table~\ref{tab:llama_extraction}, Table~\ref{tab:qwen_extraction} and Table~\ref{tab:llama_qwen_summary}.
\begin{itemize}
    \item \textbf{Table~\ref{tab:llama_extraction}} and \textbf{Table~\ref{tab:qwen_extraction}} details \textit{extraction-based} pruning results, comparing different scoring signals (\textsc{Lingua}, Entropy, NLL) across both step-level and token-level granularities for CoT Compress on LLaMA-3.1-8B-Instruct and Qwen-2.5-1.5B-Instruct.
    \item \textbf{Table~\ref{tab:llama_qwen_summary}} details \textit{generative} compression results for CoT Compress on LLaMA-3.1-8B-Instruct and Qwen-2.5-1.5B-Instruct, covering the spectrum of structural fidelity (SP, SE, SF, FR) across different prompt constraints.
\end{itemize}
For each configuration, we report the top-1 accuracy (\textbf{Acc}), the average length of the compressed CoT in tokens (\textbf{Tok}), and the end-to-end inference token ratio ($\kappa$, denoted as \textbf{InferRatio}). These metrics are reported separately for \textbf{MATH} and \textbf{GENERAL} benchmarks, covering both the \textsc{Natural Reasoning} (Long-CoT) and \textsc{LIMO} (Short-CoT) training regimes. The data points in these tables constitute the candidate set used to derive the Pareto frontiers and "knee points" discussed in the main text.

\subsection{Cross-Scale Verification on Qwen2.5-7B-Instruct}
\label{appendix:cross_scale}

To probe whether key findings hold at a larger parameter scale, we 
conduct a targeted verification on Qwen2.5-7B-Instruct under the 
Short-CoT regime, focusing on the most diagnostic conditions for 
each finding rather than exhaustively replicating the full sweep.

\begin{table}[h!]
\centering
\small
\setlength{\tabcolsep}{1.2mm}
\begin{tabular}{@{}llccccc@{}}
\toprule
\multirow{2}{*}{\textbf{Granularity}} & \multirow{2}{*}{\textbf{Signal}} 
& \multirow{2}{*}{\textbf{Ratio}} 
& \multicolumn{2}{c}{\textbf{MATH}} 
& \multicolumn{2}{c}{\textbf{GENERAL}} \\
\cmidrule(lr){4-5} \cmidrule(lr){6-7}
& & & Acc & $\rho$ & Acc & $\rho$ \\
\midrule
— & Baseline & — & 75.3 & 1.00 & 40.1 & 1.00 \\
\midrule
\multirow{3}{*}{Token} 
& Entropy   & 0.7 & 54.6 & 0.95 & 36.8 & 0.99 \\
& Entropy   & 0.5 & 46.7 & 0.97 & 37.5 & 0.86 \\
& NLL       & 0.7 & 15.8 & 1.37 & 33.5 & 1.09 \\
& LLMLingua-2 & 0.7 & 74.1 & 1.21 & 39.6 & 1.14 \\
\midrule
\multirow{2}{*}{Step}  
& Entropy   & 0.7 & 74.8 & 1.03 & 40.8 & 1.06 \\
& Entropy   & 0.5 & 73.3 & 0.87 & 40.8 & 0.91 \\
\bottomrule
\end{tabular}
\caption{Targeted extraction-based verification on Qwen2.5-7B-Instruct 
(Short-CoT). Token-level Entropy and NLL collapse on Math while 
step-level methods remain stable, consistent with findings on 
LLaMA-3.1-8B and Qwen2.5-1.5B.}
\label{tab:qwen7b_extraction}
\end{table}

\begin{table}[h!]
\centering
\small
\setlength{\tabcolsep}{1.2mm}
\begin{tabular}{@{}lcccccc@{}}
\toprule
\multirow{2}{*}{\textbf{Structure}} & \multirow{2}{*}{\textbf{Ratio}} 
& \multicolumn{2}{c}{\textbf{MATH}} 
& \multicolumn{2}{c}{\textbf{GENERAL}} \\
\cmidrule(lr){3-4} \cmidrule(lr){5-6}
& & Acc & $\rho$ & Acc & $\rho$ \\
\midrule
Baseline & —   & 75.3 & 1.00 & 40.1 & 1.00 \\
\midrule
SP       & 0.5 & 76.8 & 0.68 & 39.5 & 0.81 \\
SE       & 0.5 & 76.2 & 0.69 & 40.2 & 0.66 \\
SF       & 0.5 & 71.7 & 0.51 & 40.1 & 0.62 \\
\bottomrule
\end{tabular}
\caption{Targeted summarization-based verification on Qwen2.5-7B-Instruct 
(Short-CoT). The fidelity wall reproduces: Math accuracy degrades from 
SP to SF while General performance remains stable, consistent with 
findings across all model families.}
\label{tab:qwen7b_summary}
\end{table}


\begin{table*}[t]
\centering
\small
\setlength{\tabcolsep}{4pt}        
\renewcommand{\arraystretch}{0.95} 
\caption{Results of reflection-aligned extraction. \textbf{Lingua-LowRef} constrains Lingua to match NLL's lower reflection deletion rate, while \textbf{NLL-HighRef} forces NLL to match Lingua's higher deletion rate. Accuracy remains stable despite significant shifts in reflection retention.}
\label{tab:reflection_align}
\begin{tabular}{clcccc}
\toprule
\multirow{2}{*}{Ratio ($c$)} & \multirow{2}{*}{Method} & \multicolumn{2}{c}{Math} & \multicolumn{2}{c}{General} \\
\cmidrule(lr){3-4} \cmidrule(lr){5-6}
 & & Acc (\%) & Tok & Acc (\%) & Tok \\
\midrule
\multirow{4}{*}{0.5}
 & \textsc{Lingua-Base}   & 27.2 & 2173 & 39.9 & 1246 \\
 & \textsc{Lingua-LowRef} & 27.0 & 2315 & 38.1 & 1176 \\
 & \textsc{NLL-Base}      & 27.1 & 2319 & 42.0 & 1201 \\
 & \textsc{NLL-HighRef}   & 27.7 & 2555 & 40.5 & 1210 \\
\midrule
\multirow{4}{*}{0.7}
 & \textsc{Lingua-Base}   & 23.8 & 2575 & 38.2 & 1244 \\
 & \textsc{Lingua-LowRef} & 25.4 & 2082 & 35.9 & 1121 \\
 & \textsc{NLL-Base}      & 22.0 & 2391 & 40.5 & 1127 \\
 & \textsc{NLL-HighRef}   & 23.3 & 2474 & 38.8 & 1208 \\
\bottomrule
\end{tabular}
\end{table*}

\section{Reflection-Aligned Step Pruning}
\label{app:ref_align}

To verify whether the compressibility of Long-CoT stems from specifically targeting reflection steps, we designed a controlled experiment to decouple \textit{reflection deletion rate} from the underlying \textit{scoring signal}.

\subsection{Reflection-step density.}
\label{sec: density}
For each CoT trace $t$ with $T$ steps, we define the reflection-step density as:
\begin{equation}
\text{refl\_density}(t) = \frac{\#\text{reflection steps in }t}{T}
\end{equation}
where reflection steps are identified by a rule-based tagger over discourse markers
(e.g., “wait”, “hmm”) and self-referential commentary~\citep{jia2025epic}.

\subsection{Experimental Setup and Definitions}
For a teacher trace $\mathcal{S}=\{s_i\}_{i=1}^{N}$, each step receives a strictly defined importance score $q_i$ (via \textsc{Lingua} or \textsc{NLL}). We identify reflection steps using a fixed, case-insensitive keyword-based regular expression with word-boundary matching (e.g., \texttt{wait}, \texttt{hmm}, \texttt{let me think}, \texttt{reconsider}, \texttt{wrong}, \texttt{validate})~\citep{jia2025epic}. Formally, we define $r_i=\mathbbm{1}[\mathrm{RegexMatch}(s_i)]$, where $\mathrm{RegexMatch}(\cdot)$ returns $1$ iff any keyword is matched, and $0$ otherwise.

Given a step-level compression target $c\in\{0.5,0.7\}$, we define the step deletion ratio as $r=1-c$ and set the deletion budget as
\[
K=\mathrm{round}(r\cdot N)=\mathrm{round}((1-c)\cdot N).
\]
The natural tendency of a method to remove reflections is measured by the reflection deletion ratio:
\begin{equation}
\alpha_{\text{del}} = \frac{1}{K} \sum_{i \in \mathcal{D}} r_i,
\end{equation}
where $\mathcal{D}$ is the set of deleted step indices (with $|\mathcal{D}|=K$). Empirically, \textsc{Lingua} tends to yield a higher $\alpha_{\text{del}}$ (aggressive reflection removal) compared to \textsc{NLL}.

\subsection{Constructing Reflection-Aligned Variants}
We construct aligned variants to enforce a specific target ratio $\alpha_{\text{target}}$ while preserving the ranking logic of the base signal. We partition indices into Reflection ($\mathcal{R}$) and Other ($\mathcal{O}$) buckets.

We then allocate the deletion budget across the two buckets:
\begin{equation}
K_R = \text{round}(\alpha_{\text{target}} \cdot K), \qquad K_O = K - K_R,
\end{equation}
subject to feasibility (if one bucket is insufficient, the remainder is reassigned to the other bucket).
Within each bucket, deletions preferentially target low-score steps under the same base scoring signal, and the final selection enforces the total budget $K$.

We define two anchors based on the natural behavior of the base methods at $c \in \{0.5, 0.7\}$: $\alpha_{\text{low}}$ (from \textsc{NLL}) and $\alpha_{\text{high}}$ (from \textsc{Lingua}). This yields cross-aligned datasets:
\begin{itemize}[leftmargin=*, itemsep=0pt]
\item \textbf{NLL-HighRef}: Forces \textsc{NLL} to match \textsc{Lingua}'s aggressive reflection deletion ($\alpha_{\text{high}}$).
\item \textbf{Lingua-LowRef}: Forces \textsc{Lingua} to match \textsc{NLL}'s conservative reflection deletion ($\alpha_{\text{low}}$).
\end{itemize}

\subsection{Results: Redundancy is Diffuse}
We trained students on these variants and compared them to unconstrained baselines. As shown in Table~\ref{tab:reflection_align}, \textbf{the impact of reflection alignment was negligible.} At matched step budgets ($K$), manipulating $\alpha_{\text{del}}$ produced no significant deviation in downstream accuracy. Specifically:
\begin{itemize}
\item Forcing \textsc{NLL} to delete more reflections did not improve its compression efficiency.
\item Restricting \textsc{Lingua} to keep more reflections did not degrade its performance.
\end{itemize}
These results confirm that the "safe regime" in Long-CoT is not driven by the specific removal of reflection steps. Instead, redundancy appears to be diffuse: critical information is distributed across both reflective and non-reflective steps, and the model is robust to pruning as long as the global reasoning skeleton is maintained.

\section{Detailed Analysis of Token Preservation}
\label{app:token_analysis}

\begin{table*}[t]
\centering
\small
\begin{tabular}{lccccccccc}
\toprule
& \multicolumn{3}{c}{0.1} & \multicolumn{3}{c}{0.3} & \multicolumn{3}{c}{0.5} \\
Signal & Sym & Stop & Other & Sym & Stop & Other & Sym & Stop & Other \\
\midrule
Entropy   & 0.10 & 0.11 & 0.08 & 0.30 & 0.29 & 0.21 & 0.49 & 0.44 & 0.36 \\
NLL       & 0.11 & 0.12 & 0.07 & 0.30 & 0.28 & 0.21 & 0.51 & 0.40 & 0.34 \\
LLMLingua & 0.03 & 0.29 & 0.02 & 0.15 & 0.54 & 0.11 & 0.32 & 0.68 & 0.25 \\
\bottomrule
\end{tabular}
\caption{Extended breakdown of deleted token types on \textsc{NaturalReasoning} for token-level pruning across target compression ratios.}
\label{tab:token_deletion_types_full}
\end{table*}

To understand the qualitative behavior of different pruning signals, we categorize the tokens removed by each method. We implemented a Python-based classifier to map every deleted token into one of three mutually exclusive categories based on the heuristics defined below.

\subsection{Token Categorization Methodology}
We first tokenize text via whitespace and strip trailing punctuation. Each token is then classified sequentially according to the following priority rules, strictly aligning with our evaluation script:

\begin{enumerate}[leftmargin=*, itemsep=2pt]
    \item \textbf{Class 1: Constraint-Bearing \& Symbolic (High Priority).}
    This class captures the structural anchors of the reasoning chain. It includes:
    \begin{itemize}
        \item \textbf{Logic Markers:} Key deductive connectives (e.g., \textit{"let", "assume", "implies", "therefore", "since", "if", "iff"}).
        \item \textbf{Math Operators \& Symbols:} All standard arithmetic operators, comparators, and Unicode mathematical symbols (e.g., $\in, \subseteq, \forall, \sum$).
        \item \textbf{\LaTeX{} Commands:} Substrings indicative of formal notation (e.g., \texttt{\textbackslash frac}, \texttt{\textbackslash boxed}, \texttt{\textbackslash sqrt}).
        \item \textbf{Numerical Patterns:} Any token containing digits, including scientific notation and fractions.
    \end{itemize}

    \item \textbf{Class 2: Stopwords.}
    Tokens found in standard English stopword lists (sourced from NLTK~\citep{bird-loper-2004-nltk}/scikit-learn).
    \textbf{Crucially}, to prevent misclassification, any stopword that also appears in the \textit{Logic Markers} list (e.g., "if", "not") is removed from this set and forced into Class 1. Thus, Class 2 strictly represents low-semantic syntactic glue.

    \item \textbf{Class 3: Content.}
    All remaining tokens that do not fall into Class 1 or Class 2. These are typically domain-specific nouns, verbs, and adjectives describing the problem context.
\end{enumerate}

\subsection{Results: Deletion Distribution}
Table~\ref{tab:token_deletion_types_full} presents the \textbf{Deletion Rate} for each category, defined as the fraction of tokens of that type removed from the original trace.

The data reveals a clear behavioral signature: \textsc{LLMLingua-2} acts as a "logic-aware" filter. Even at aggressive compression levels, it consistently yields significantly lower deletion rates for \textbf{Sym} (Class 1) tokens compared to Entropy and NLL baselines, while shifting the deletion budget heavily towards \textbf{Stop} (Class 2) tokens. This confirms that semantic probability estimators are better aligned with the logical structure of chain-of-thought reasoning than simple perplexity-based metrics.

\subsection{Validation of the Symbolic Anchor Hypothesis}
\label{app:math_protection_ablation}

To further verify that loss of symbolic structure in NLL-based degradation, we conducted a controlled experiment with a restricted pruning scope.
In this setting, we excluded tokens categorized as math-like patterns from the pruning candidate set entirely. 
Consequently, the NLL-based ranking and deletion were applied solely to the remaining vocabulary (i.e., stopwords and other content tokens), while all math-related tokens were preserved by default.

As shown in Table~\ref{tab:nll_protection_ablation}, this selective retention strategy significantly improved the stability of NLL-based pruning.
By simply preventing the removal of these symbolic anchors, the optimized NLL baseline recovered a large portion of the performance gap relative to LLMLingua2.
This result confirms that the standard NLL metric fails in high-risk settings specifically because it assigns low probabilities to essential symbolic operators, not because the metric itself is incapable of ranking redundant prose.

\begin{table}[t]
\centering
\small 
\setlength{\tabcolsep}{4pt} 
\begin{tabular}{l cc cc}
\toprule
& \multicolumn{2}{c}{\textbf{Math}} & \multicolumn{2}{c}{\textbf{General}} \\
\cmidrule(lr){2-3} \cmidrule(lr){4-5}
\textbf{Ratio} & NLL & +Protect & NLL & +Protect \\
\midrule
0.9 & 0.632 & \textbf{0.675} & 0.360 & \textbf{0.364} \\
0.7 & 0.422 & \textbf{0.632} & 0.367 & \textbf{0.376} \\
0.5 & 0.333 & \textbf{0.608} & 0.348 & \textbf{0.404} \\
\bottomrule
\end{tabular}
\caption{Ablation study on token protection during NLL pruning. Results reported on Llama 3.1 8B Instruct (Short CoT). ``+Protect'' denotes NLL pruning where math-related tokens are excluded from the deletion candidates.}
\label{tab:nll_protection_ablation}
\end{table}
\section{Step-Level Decision Overlap Analysis}
\label{app:step_overlap}

\begin{table}[t]
\centering
\begin{tabular}{lccc}
\toprule
Pair & 0.1 & 0.3 & 0.5 \\
\midrule
ENT--L2  & 0.09 & 0.16 & 0.31 \\
NLL--L2  & 0.10 & 0.17 & 0.33 \\
ENT--NLL & 0.70 & 0.70 & 0.76 \\
\bottomrule
\end{tabular}
\caption{Pairwise Jaccard overlap between \textbf{deleted} step sets for different step-level scoring signals under matched target ratios. (ENT: Entropy; L2: LLMLingua2).}
\label{tab:jaccard_deleted}
\end{table}

To empirically verify the claim that different pruning signals converge on a shared reasoning backbone while disagreeing on peripheral redundancy, we analyzed the pairwise consensus of their pruning decisions.

\begin{table*}[t]
\centering
\small
\setlength{\tabcolsep}{5pt}
\renewcommand{\arraystretch}{1.1}
\caption{
\textbf{Coherence Evaluation Results.} Pairwise comparison of semantic compression (\textsc{LLMLingua-2}) versus naive ratio pruning over 100 samples per setting. ``Math-Prot'' denotes the variant with protection for mathematical symbols.
}
\label{tab:coherence}
\begin{tabular}{llcrrr}
\toprule
\textbf{Granularity} & \textbf{Configuration} & \textbf{Target} $\rho$ & \textbf{A Wins} & \textbf{B Wins} & \textbf{Win Rate (A)} \\
\midrule
Step & LLMLingua-2 & 0.7 & 95 & 3 & 0.960 \\
Step & LLMLingua-2 & 0.9 & 92 & 5 & 0.935 \\
\midrule
Token & LLMLingua-2 & 0.7 & 100 & 0 & 1.000 \\
Token & LLMLingua-2 & 0.9 & 100 & 0 & 1.000 \\
Token & LLMLingua-2 (Math-Prot) & 0.7 & 100 & 0 & 1.000 \\
\bottomrule
\end{tabular}
\end{table*}

\subsection{Calculation Methodology}
We implemented a strict index-based comparison pipeline. For any pair of pruning methods $\mathcal{M}_A$ and $\mathcal{M}_B$ (e.g., Entropy vs. LLMLingua-2):

\begin{enumerate}[leftmargin=*, itemsep=2pt]
    \item \textbf{Alignment:} We align samples by question ID across the generated `.jsonl` outputs.
    \item \textbf{Index Mapping:} For each sample, we map the retained text segments back to the original teacher step indices $\{1, \dots, N\}$.
    \item \textbf{Set Construction:} We derive the set of \textit{deleted} indices for each method, denoted as $\mathcal{D}_A$ and $\mathcal{D}_B$.
    \item \textbf{Metric:} We compute the Jaccard Similarity Coefficient for the deletion sets:
    \begin{equation}
        J(\mathcal{D}_A, \mathcal{D}_B) = \frac{|\mathcal{D}_A \cap \mathcal{D}_B|}{|\mathcal{D}_A \cup \mathcal{D}_B|}
    \end{equation}
\end{enumerate}
This process is repeated for each target ratio and averaged across the dataset.

\subsection{Divergence in Deletion Choices}
Table~\ref{tab:jaccard_deleted} presents the overlap of \textbf{deleted steps} across methods. We observe a clear dichotomy:
\begin{itemize}
    \item \textbf{High Internal Consistency:} Entropy (ENT) and NLL show high overlap ($>0.70$), indicating that probability-based metrics largely identify the same "easy" tokens to remove.
    \item \textbf{Low Cross-Paradigm Overlap:} When comparing probability-based methods against the symbol-aware LLMLingua-2 (L2), the overlap drops significantly (0.10--0.33).
\end{itemize}

Despite this substantial disagreement on \textit{what to delete}, the main text (Section~\ref{sec: Singal}) shows that their downstream accuracy is nearly identical. This confirms that the non-overlapping deleted steps represent \textbf{interchangeable redundancy}—peripheral information whose removal (whether by probability or semantic estimation) does not disrupt the core reasoning chain.


\section{LLM-Based Coherence Judge}
\label{app:coherence}

While quantitative metrics measure task accuracy, they do not fully capture the linguistic quality of the compressed traces. Our analysis suggests that token-level pruning—unlike step-level selection—can severely disrupt local continuity and readability. While other method~\citep{wang2025r1compress} measures a token-level LM loss as a proxy of inter-chunk coherence, we are interested in how coherent a compressed CoT is as a \emph{teaching script} for smaller models, so we implement an LLM-based coherence judge to perform pairwise comparisons.

\paragraph{Task Definition.}
We evaluate the \textit{reasoning coherence} of two Chain-of-Thought (CoT) rationales (Method A vs. Method B) generated for the same question. The judge is explicitly instructed to ignore the correctness of the final answer and focus solely on linguistic and logical flow. The output is a decision label in $\{\texttt{A}, \texttt{B}, \texttt{TIE}\}$.

\paragraph{Judge Implementation.}
We utilize a powerful LLM, \texttt{gemini-2.5-flash}~\citep{comanici2025gemini25pushingfrontier}, as the judge. The prompt instructs the model to evaluate: (i) global structural organization, (ii) logical continuity between steps, (iii) proper usage of intermediate conclusions, and (iv) the absence of disjointed or "broken" text.

\begin{tcolorbox}[
colback=blue!5!white,
        colframe=blue!75!black,
        title=The pairwise comparison prompt used for the coherence judge.,
        fonttitle=\small\bfseries, 
        fontupper=\footnotesize,
        top=2mm, 
        bottom=2mm,
        left=2mm,
        right=2mm,
        boxsep=1mm, 
]
You are evaluating the quality of two chain-of-thought (CoT) rationales
for the **same question**. Your goal is to decide which CoT is more
coherent and easier to follow **as a reasoning process**, ignoring whether
the final answer is correct.

Do **not** consider whether the final numerical/choice answer is correct.
You should focus only on how clear and coherent the reasoning itself is.

[Question]
{question}

[CoT A]
$${cot_a}$$

[CoT B]
$${cot_b}$$

Please compare CoT A and CoT B only in terms of **reasoning coherence**,
according to the following criteria:

1. Global organization: the rationale states the goal/known facts clearly
   and follows a sensible overall plan.
2. Step-to-step logic: each step follows logically from previous steps,
   without unexplained jumps.
3. Use of intermediate conclusions: important intermediate results are
   clearly derived and then used later.
4. Noise and redundancy: the rationale avoids unnecessary digressions,
   off-topic commentary, or excessive repetition that makes it harder to
   follow.

If one CoT is much shorter but clearly skips important steps, treat it as
**less** coherent than a slightly longer but well-explained CoT.
Ignore minor formatting issues such as non-consecutive step numbers.
Focus on whether the reasoning itself is logically coherent and easy to follow.

If one CoT is more coherent, choose that one. If their coherence is very
similar, choose TIE.

First, briefly explain your comparison in 2–3 sentences.
Then, on a separate line, output your decision in the format:

DECISION: A
or
DECISION: B
or
DECISION: TIE
\end{tcolorbox}

\newpage
\paragraph{Aggregation.}
For each experimental configuration, we evaluate 100 pairs. The win rate for Method A is calculated as:
\[
\text{WinRate(A)} = \frac{\#(\texttt{A}) + 0.5\cdot \#(\texttt{TIE})}{\#(\texttt{A}) + \#(\texttt{B}) + \#(\texttt{TIE})}.
\]

\paragraph{Results: Extraction}
Table~\ref{tab:coherence} summarizes the pairwise coherence results comparing semantic compression (Method A: \textsc{LLMLingua-2}) against a standard length-based baseline (Method B: \textsc{Ratio}). The results overwhelmingly favor semantic compression, particularly at the token level (100\% win rate), confirming that naive token deletion catastrophic destroys coherence, whereas semantic estimators preserve readability.

\paragraph{Results: Summarization}
Table~\ref{tab:coherence_summarization} extends the evaluation to 
summarization-based methods, comparing each structural fidelity 
level against the extraction baseline across compression ratios.

\begin{table}[h!]
\centering
\small
\begin{tabular}{lcccc}
\toprule
\textbf{Structure} & \textbf{$\gamma=0.5$} & \textbf{$\gamma=0.7$} 
& \textbf{$\gamma=0.9$} & \textbf{Free} \\
\midrule
None   & 0.41 & —    & —    & —    \\
Light  & 0.52 & 0.53 & 0.73 & 0.21 \\
Medium & 0.41 & 0.58 & 0.79 & 0.30 \\
Strong & 0.12 & 0.24 & 0.34 & 0.38 \\
\bottomrule
\end{tabular}
\caption{Coherence win rates of summarization methods against an 
extraction baseline (win rate $>0.5$ favors summarization).}
\label{tab:coherence_summarization}
\end{table}

Structure Light and Medium show increasing coherence win rates as compression tightens, reaching 0.73 and 0.79 at $\gamma=0.9$. 
Structure Strong consistently scores below 0.5 across all settings, reflecting surface-level fragmentation from strict token-removal constraints---a pattern consistent with its strong Math accuracy in the main experiments. Structure None yields the lowest win rate overall, aligning with its higher output variance and degraded Math performance observed in Section~\ref{sec:Structure}.

\section{Quantitative verification of generative rewriting}
\label{app:Quantitative_verification}

\subsection{Quantitative verification of restructuring order}

\paragraph{Sample construction.}
To verify that the four restructuring levels are ordered by depth in an objective manner, we construct a frozen, balanced sample of 800 rewritten traces. The sample includes both supervision regimes: LIMO for Long-CoT and NaturalReasoning for Short-CoT, with 100 traces per restructuring level per dataset. Sampling is performed with a single fixed seed, and compressed traces are aligned to their original uncompressed traces before evaluation.

\paragraph{Metric.}
We compute the metric copy\_rate, defined as the fraction of the rewritten trace copied verbatim from the original in spans of at least five tokens. Before matching, both traces are normalized by lowercasing, removing leading fillers, and collapsing whitespace and punctuation spacing. Verbatim spans are extracted using a SequenceMatcher procedure. The metric is formally defined as
\[
\mathrm{copy\_rate} = \frac{\text{matched verbatim tokens}}{\text{total tokens in rewritten trace}}.
\]
Higher values indicate that the rewritten trace remains closer to the original wording, whereas lower values indicate deeper restructuring.

\paragraph{Results.}
The measured copy\_rate decreases monotonically from L1 to L4 in both datasets. For LIMO, the median copy\_rate values are approximately $0.664$, $0.187$, $0.096$, and $0.035$ for L1--L4, respectively; for NaturalReasoning, the corresponding values are approximately $0.635$, $0.234$, $0.120$, and $0.041$. We also compute the Spearman rank correlation between restructuring level and copy\_rate; the correlation is negative and highly significant in both datasets: for LIMO, $\rho=-0.6757$ with $p=1.14\times10^{-54}$; for NaturalReasoning, $\rho=-0.6454$ with $p=1.70\times10^{-48}$. These results show that the intended ordering of restructuring depth is recovered objectively and is not merely assumed from prompt design or model-specific behavior.

\begin{table}[t]
\centering
\resizebox{\columnwidth}{!}{%
\begin{tabular}{lcc}
\toprule
\shortstack{\textbf{Restructuring}\\\textbf{Level}} &
\shortstack{\textbf{LIMO}\\\textbf{Median [Q1, Q3]}} &
\shortstack{\textbf{NaturalReasoning}\\\textbf{Median [Q1, Q3]}} \\
\midrule
L1 SP
& 0.664 [0.364, 0.910]
& 0.635 [0.419, 0.866] \\

L2 SE
& 0.187 [0.116, 0.190]
& 0.234 [0.205, 0.325] \\

L3 SF
& 0.096 [0.042, 0.133]
& 0.120 [0.115, 0.140] \\

L4 FR
& 0.035 [0.022, 0.049]
& 0.041 [0.018, 0.067] \\

\midrule
Spearman $\rho$
& $-0.68$, $p < 1\mathrm{e}{-54}$
& $-0.65$, $p < 1\mathrm{e}{-48}$ \\
\bottomrule
\end{tabular}%
}
\caption{Copy-rate summary by dataset and restructuring level.}
\label{tab:copy_rate_summary}
\end{table}

\subsection{Quantitative verification of answer preservation}

\paragraph{Evaluation setup.}
We further verify that increasing restructuring level does not systematically alter the final answer. To this end, answer preservation was evaluated using a blind judge with DeepSeek-V4-Flash~\citep{deepseekai2026deepseekv4} at temperature $0$. The judge is asked whether the rewritten trace reaches the same final answer as the original trace, and the restructuring level is not provided to the judge. The judge output is a structured response with values \texttt{true}, \texttt{false}, or \texttt{NA}, and only comparable items are included in the denominator.

\paragraph{Metric.}
The answer preservation rate measures the fraction of comparable rewritten
traces that produce the same final answer as the original trace.

\paragraph{Results.}
Across all levels, answer preservation remains high in both datasets. For LIMO, the preservation rates are $94\%$, $91\%$, $87\%$, and $86\%$ for L1--L4, respectively; for NaturalReasoning, the rates are $99\%$, $91\%$, $98\%$, and $93\%$. These results indicate that stronger rewriting does not systematically break the final answer, even when the rewritten trace is substantially more compressed or structurally transformed.

\begin{table}[t]
\centering
\resizebox{\columnwidth}{!}{%
\begin{tabular}{lcc}
\toprule
\textbf{Restructuring level} & \textbf{LIMO} & \textbf{NaturalReasoning} \\
\midrule
L1 Structural Preservation & 0.94 & 0.99 \\
L2 Structural Editing      & 0.91 & 0.91 \\
L3 Semantic Fusion         & 0.87 & 0.98 \\
L4 Framework Replacement   & 0.86 & 0.93 \\
\midrule
Overall                    & 0.895 & 0.952 \\
\bottomrule
\end{tabular}%
}
\caption{Answer-preservation rates by dataset and restructuring level.}
\label{tab:answer_preservation_summary}
\end{table}
\clearpage
\onecolumn
\section{CoT Compression Case Study}
\label{appendix:compress_cases}
\definecolor{LightGray}{gray}{0.95}
\definecolor{LightCyan}{rgb}{0.88,1,1}

\begingroup
\renewcommand{\arraystretch}{1.4} 

\begin{table}[h!]
\centering
\setlength{\tabcolsep}{8pt} 
\fontsize{8}{10}\selectfont 

\begin{tabular}{p{0.12\linewidth} p{0.78\linewidth} r} 
\toprule

\rowcolor{white}
\textbf{question} & \textbf{Calculate the energy needed for a 1 kg mass to accelerate from 2 m/s to 6 m/s, considering the reference frame for kinetic energy and speed. Provide a detailed explanation of your calculation, including how you account for the change in kinetic energy and the relevance of the reference frame.} \\
\midrule

\rowcolor{LightGray}
\textbf{Extraction Method} & \textbf{Generated Output} \\ 
\midrule

\rowcolor{white}
\textbf{Original} & \#\# Step 1: Identify the formula for kinetic energy
The kinetic energy ($KE$) of an object is given by the formula $KE = 0.5 * m * v^2$, where $m$ is the mass of the object and $v$ is its velocity.

\#\# Step 2: Calculate the initial kinetic energy
To find the initial kinetic energy, we use the given initial velocity ($v_1 = 2$ m/s) and the mass ($m = 1$ kg) in the kinetic energy formula: $KE_1 = 0.5 * 1 * (2)^2 = 0.5 * 1 * 4 = 2$ Joules.

\#\# Step 3: Calculate the final kinetic energy
Using the given final velocity ($v_2 = 6$ m/s) and the same mass ($m = 1$ kg) in the kinetic energy formula: $KE_2 = 0.5 * 1 * (6)^2 = 0.5 * 1 * 36 = 18$ Joules.

\#\# Step 4: Determine the change in kinetic energy
The change in kinetic energy ($\Delta KE$) is the difference between the final and initial kinetic energies: $\Delta KE = KE_2 - KE_1 = 18 - 2 = 16$ Joules.

\#\# Step 5: Consider the relevance of the reference frame
The calculation of kinetic energy and the change in kinetic energy depends on the reference frame, as kinetic energy is a frame-dependent quantity. However, since we are calculating the change in kinetic energy for an object accelerating from one speed to another within the same reference frame (implied by the problem statement), the reference frame does not affect the calculation of the energy needed for the acceleration in this context. The energy needed is the same regardless of the observer's frame of reference because the initial and final velocities are given relative to the same frame.

The final answer is: $\boxed{16}$ \\

\rowcolor{gray!30}
\textbf{Word level Entropy} at 50\% compression ratio & \#\# Step Identify formula kinetic
The kinetic (KE) of is $KE 0.5 * m$ and velocity.

\#\# Calculate initial kinetic
To find initial energy, we use given initial velocity (v1 = m/s) and the mass (m kg) kinetic formula: $KE1 1 * (2)^2 = 0.5$ Joules.

\#\# Calculate
Using given the same (m kg) formula: Determine change
The ($\Delta KE$) is the difference final and energies: - Consider the relevance the
The calculation of kinetic and the change in kinetic depends on the reference frame, as kinetic is a frame-dependent quantity. However, since we are calculating the change kinetic for an object accelerating from one speed within the reference (implied by the problem statement), the reference frame does not affect the calculation of the energy needed for the acceleration in this context. The energy needed is the same regardless of the observer's frame of reference because the initial and final velocities are given relative to the same frame.

The final answer is: $\boxed{16}$ \\

\rowcolor{white}
\textbf{Word level LLMLingua2} at 50\% compression ratio & Step Identify formula kinetic energy formula $KE = 0.5 * m * v^2$ $m$ mass $v$ velocity

Calculate initial kinetic energy velocity ($v_1 = 2$ m/s) mass ($m = 1$ kg) $KE_1 = 0.5 * 1 * (2)^2 = 0.5 * 1 * 4 = 2$ Joules

Calculate final kinetic energy velocity ($v_2 = 6$ m/s) mass ($m = 1$ kg) $KE_2 = 0.5 * 1 * (6)^2 = 0.5 * 1 * 36 = 18$ Joules

Determine change in kinetic energy difference final initial kinetic energies $\Delta KE = KE_2 - KE_1 = 18 - 2 = 16$ Joules

Consider relevance reference frame calculation kinetic energy depends reference frame accelerating another within the same reference frame (implied by the problem statement), the reference frame does not affect the calculation of the energy needed for the acceleration in this context. The energy needed is the same regardless of the observer's frame of reference because the initial and final velocities are given relative to the same frame.

The final answer is: $\boxed{16}$ \\

\rowcolor{gray!30}
\textbf{Word-level NLL} at 50\% compression ratio & \#\# Step Identify formula kinetic
The kinetic (KE) of is given $KE 0.5 * m$ of and its velocity.

\#\# Calculate initial kinetic
To find initial energy, we use given initial velocity ($v_1 = $ m/s) and the mass ($m$ kg) kinetic formula: $KE_1 1 * (2)^2 = 0.5$ Joules.

\#\# Calculate
Using given same kg) formula: Determine change
The ($\Delta KE$) is the difference and initial energies: - Joules.

\#\# Consider the relevance
The calculation of kinetic and the change kinetic depends on reference frame, as kinetic is a frame-dependent quantity. However, since we are calculating the change for an object accelerating from one speed within the same reference (implied by the problem statement), the reference does affect the calculation of the energy needed for the acceleration in this context. The energy needed is the same regardless of the observer's frame of reference because the initial and final velocities are given relative to the same frame.

The final answer is: $\boxed{16}$ \\

\bottomrule
\end{tabular}
\caption{Case study: reflection-aligned extraction at 50\% compression. For the same kinetic-energy prompt, we report the original uncompressed response and extracted outputs produced using token-level redundancy signals (Entropy, Lingua, and NLL).} 

\end{table}
\endgroup

\definecolor{LightGray}{gray}{0.95}
\definecolor{LightCyan}{rgb}{0.88,1,1}

\begingroup
\renewcommand{\arraystretch}{1.4} 

\begin{table}[h!]
\centering
\setlength{\tabcolsep}{8pt} 
\fontsize{8}{10}\selectfont 

\begin{tabular}{p{0.12\linewidth} p{0.78\linewidth} r} 
\toprule
\rowcolor{white}
\textbf{question} & \textbf{Calculate the energy needed for a 1 kg mass to accelerate from 2 m/s to 6 m/s, considering the reference frame for kinetic energy and speed. Provide a detailed explanation of your calculation, including how you account for the change in kinetic energy and the relevance of the reference frame.} \\
\midrule
\rowcolor{LightGray}
\textbf{Generative Rewriting} & \textbf{Generated Output} \\ 
\midrule

\rowcolor{white}
\textbf{L4}:\textbf{Framework Replacement} at 50\% compression ratio& **Analysis:** Determine energy for 1\,kg accelerating 2\,m/s to 6\,m/s ($\Delta KE$). Critically, address reference frame impact on $KE$ and required energy.

**Initial Approach:** Compute initial $KE$ ($0.5*1*2^2$) and final $KE$ ($0.5*1*6^2$). Energy needed is $\Delta KE$. $KE$ depends on frame.

**Reflection:** Though $KE$ is frame-dependent, *work done* ($\Delta KE$) is invariant across inertial frames if velocities are consistently measured from one point.

**Improvement:** $KE_1 = 2$ J; $KE_2 = 18$ J. So, $\Delta KE = 16$ J. This energy change remains consistent for all inertial observers for this specific acceleration.

**Summary:** Energy needed = $\Delta KE = 16$ J (18 J final - 2 J initial). This change is invariant across inertial frames, as velocities were consistently defined.

ANSWER: $\boxed{16}$ \\

\rowcolor{gray!30}
\textbf{L3}:\textbf{Semantic Fusion} at 50\% compression ratio & The energy needed is the change in kinetic energy. First, calculate the initial kinetic energy ($KE_1$) using $KE = 0.5 * m * v^2$: $KE_1 = 0.5 * 1\,\text{kg} * (2\,\text{m/s})^2 = 2$ Joules. Next, calculate the final kinetic energy ($KE_2$): $KE_2 = 0.5 * 1\,\text{kg} * (6\,\text{m/s})^2 = 18$ Joules. The energy required is $\Delta KE = KE_2 - KE_1 = 18\,\text{J} - 2\,\text{J} = 16$ Joules. Although kinetic energy is frame-dependent, the change in kinetic energy for acceleration from 2 m/s to 6 m/s is calculated within the same reference frame, making the energy needed ($16$ Joules) a consistent and relevant value for this specific scenario. \\

\rowcolor{white}
\textbf{L2}:\textbf{Structural Editing} at 50\% compression ratio & The energy required for acceleration is equivalent to the change in the object's kinetic energy ($\Delta KE$). Kinetic energy ($KE$) is calculated using the formula $KE = 0.5 * m * v^2$.

For the 1 kg mass:
1.  Initial Kinetic Energy: At 2 m/s, $KE_1 = 0.5 * 1\,\text{kg} * (2\,\text{m/s})^2 = 2$ J.
2.  Final Kinetic Energy: At 6 m/s, $KE_2 = 0.5 * 1\,\text{kg} * (6\,\text{m/s})^2 = 18$ J.

The energy needed is the difference: $\Delta KE = KE_2 - KE_1 = 18\,\text{J} - 2\,\text{J} = 16$ J.

While kinetic energy itself is frame-dependent, the *change* in kinetic energy and the associated energy input are frame-independent when the initial and final velocities are measured within the same reference frame, as specified in this problem. Therefore, the calculated 16 J accurately reflects the energy needed regardless of the chosen inertial reference frame.

ANSWER: $\boxed{16}$ \\

\rowcolor{gray!30}
\textbf{L1}:\textbf{Structural Preservation} at 50\% compression ratio & \#\# Step 1: Identify the formula for kinetic energy
Kinetic energy ($KE$) formula: $KE = 0.5 * m * v^2$; $m$ is mass, $v$ is velocity.

\#\# Step 2: Calculate the initial kinetic energy
Initial KE: Use $v_1 = 2$ m/s, $m = 1$ kg in KE formula: $KE_1 = 0.5 * 1 * (2)^2 = 0.5 * 1 * 4 = 2$ Joules.

\#\# Step 3: Calculate the final kinetic energy
Final KE: Use $v_2 = 6$ m/s, $m = 1$ kg in KE formula: $KE_2 = 0.5 * 1 * (6)^2 = 0.5 * 1 * 36 = 18$ Joules.

\#\# Step 4: Determine the change in kinetic energy
Change in KE ($\Delta KE$), difference between final, initial KEs: $\Delta KE = KE_2 - KE_1 = 18 - 2 = 16$ Joules.

\#\# Step 5: Consider the relevance of the reference frame
Calculation of $KE$, $\Delta KE$ depends on reference frame; KE is frame-dependent. However, calculating $\Delta KE$ for an object accelerating within same reference frame (implied), frame does not affect energy needed. Energy needed is same regardless of observer's frame because initial, final velocities are relative to same frame.

ANSWER: $\boxed{16}$ \\

\bottomrule
\end{tabular}
\caption{Case study: generative rewriting at 50\% compression. For the same prompt, we report summaries generated without a redundancy signal and under different structure constraints (Structure Light/Medium/Strong), illustrating the effect of format control.} 

\end{table}
\endgroup
\clearpage
\onecolumn
\section{Inference-Side Failure Cases}
\label{appendix:failure_cases}

\definecolor{LightGray}{gray}{0.95}

\begingroup
\renewcommand{\arraystretch}{1.4}

\begin{table}[h!]
\centering
\setlength{\tabcolsep}{8pt}
\fontsize{8}{10}\selectfont

\begin{tabular}{p{0.12\linewidth} p{0.78\linewidth}}
\toprule

\rowcolor{white}
\textbf{question} & \textbf{If $f(x) = \frac{3x-2}{x-2}$, find $f(-2)+f(-1)+f(0)$.} \\
\midrule

\rowcolor{LightGray}
\textbf{Method} & \textbf{Generated Output} \\
\midrule

\rowcolor{white}
Structural Preservation (\checkmark) &
$\cdots$ Step 4: $f(-2)=\frac{-8}{-4}=2$. Step 5: $f(-1)=\frac{-5}{-3}=\frac{5}{3}$. Step 6: $f(0)=\frac{-2}{-2}=1$. $\cdots$ Step 9: $\frac{9}{3}+\frac{5}{3}=\frac{14}{3}$. \textbf{ANSWER:} $\boxed{\frac{14}{3}}$ \\

\rowcolor{gray!30}
Framework Replacement (\ding{55}) &
\textbf{Analysis:} The problem requires calculating $f(-2)+f(-1)+f(0)$ for $f(x)=\frac{3x-2}{x-2}$. The expression is a rational function, meaning it is undefined at $x=2$. This means $f(-2)$, $f(-1)$, $f(0)$ are undefined. Since the sum of undefined values is also undefined, the final answer is undefined. \textbf{Summary:} The function $f(x)$ is undefined for $x=2$. Their sum is also undefined. \textbf{Final Answer:} $\boxed{\text{undefined}}$ \\

\bottomrule
\end{tabular}
\caption{Case 1: The Structure None student bypasses actual substitution by latching onto a superficially relevant concept---the undefined point of the function---a direct consequence of losing step-level computational scaffolding during training.}
\end{table}

\begin{table}[h!]
\centering
\setlength{\tabcolsep}{8pt}
\fontsize{8}{10}\selectfont

\begin{tabular}{p{0.12\linewidth} p{0.78\linewidth}}
\toprule

\rowcolor{white}
\textbf{question} & \textbf{The solution to $-4 < 2(x-1) < 8$ is expressed in the form $a < x < b$. Find the value of $a+b$.} \\
\midrule

\rowcolor{LightGray}
\textbf{Method} & \textbf{Generated Output} \\
\midrule

\rowcolor{white}
LLMLingua-2 word-level pruning (\checkmark) &
Step 1: Solve inequality $-4 < 2(x-1) < 8$ $\cdots$ $-2 < x-1 < 4$, add 1 to all sides inequality $-2+1 < x-1+1 < 4+1$. Step 2: Simplify inequalities simplify $-1 < x < 5$. Step 3: Determine values $a$ and $b$: $a=-1$, $b=5$ from inequality $-1 < x < 5$. Step 4: Calculate $a+b$: $a+b=-1+5=4$. \textbf{The final answer is:} $\boxed{4}$ \\

\rowcolor{gray!30}
NLL word-level pruning (\ding{55}) &
Step 1: Solve for $x$ in the inequality $-4 < 2(x1) < 8$. $\cdots$ First, we have $-4 < 2(x1)$, then we have $2(x1) < 8$. $\cdots$ Next, solve the second part of the inequality $2(x1) < 8$. Dividing both sides by 2, we get $x < 8$. Combine the solutions of the two inequalities. The combined solution is $-1 < x < 8$. $\cdots$ Calculate $a+b$. $-1+ = 7$. \textbf{The final answer is:} $\boxed{7}$ \\

\bottomrule
\end{tabular}
\caption{Case 2: Token-Level Signal Sensitivity. As shown above, NLL-based pruning removes constraint-bearing tokens at substantially higher rates than LLMLingua-2. This propagates directly to inference: the NLL student drops the critical key symbols in $2(x-1)<8$, leading to final calculation errors, while the LLMLingua-2 student preserves all key numerical steps despite similarly fragmented prose.}
\end{table}

\endgroup

\end{document}